\documentclass[fleqn,usenatbib]{mnras}

\usepackage{newtxtext,newtxmath}
\usepackage{multirow} 
\usepackage{booktabs}
\usepackage{xcolor}
\usepackage{array}
\usepackage{tabularx}
\usepackage{amsmath}
\usepackage{threeparttable}
\usepackage{natbib}
\usepackage{caption}
\usepackage{CJK}
\usepackage{array}
\newcolumntype{Y}{>{\raggedright\arraybackslash}X}		

\usepackage[T1]{fontenc}

\DeclareRobustCommand{\VAN}[3]{#2}
\let\VANthebibliography\thebibliography
\def\thebibliography{\DeclareRobustCommand{\VAN}[3]{##3}\VANthebibliography}

\usepackage{graphicx}	
\usepackage{amsmath}	

\title[Solar flare prediction { with CDR framework}]{Prediction of Major Solar Flares Using Interpretable { Class-dependent Reward Framework } with Active Region Magnetograms and Domain Knowledge}

\author[]{
	Zixian Wu,$^{1,2}$
	Xuebao Li,$^{1,2}$\thanks{E-mail: 305122880@qq.com}
	Yanfang Zheng,$^{1,2}$\thanks{E-mail:  zyf062856@163.com}
	Rui Wang,$^{2}$
	Shunhuang Zhang,$^{1}$
	Jinfang Wei,$^{3,1}$
	Yongshang Lv,$^{1}$
	\newauthor
	Liang Dong,$^{4}$
	Zamri Zainal Abidin,$^{5,6}$
	Noraisyah Mohamed Shah,$^{7}$
	Hongwei Ye,$^{1}$
	Pengchao Yan,$^{1}$
	Xuefeng Li,$^{1}$
	\newauthor
	Xiaojia Ji,$^{1}$
	Xusheng Huang,$^{1}$
	Xiaotian Wang,$^{1}$
	and Honglei Jin$^{1}$
	\\
	$^{1}$School of Computer Science, Jiangsu University of Science and Technology, Zhenjiang 212100, People$^{\prime}$s Republic of China\\
	$^{2}$State Key Laboratory of Space Weather, Chinese Academy of Sciences, Beijing 100190, People$^{\prime}$s Republic of China\\
	$^{3}$School of software, Southeast University, Nanjing, People$^{\prime}$s Republic of China\\
	$^{4}$Yunnan Astronomical Observatory, Chinese Academy of Sciences, Kunming 650216, People$^{\prime}$s Republic of China\\
	$^{5}$Radio Cosmology Lab, Centre for Astronomy and Astrophysics Research, Department of Physics, Faculty of Science, Universiti Malaya, 50603 Kuala Lumpur, Malaysia\\
	$^{6}$National Centre for Particle Physics, Universiti Malaya, 50603 Kuala Lumpur, Malaysia\\
	$^{7}$Department of Electrical Engineering, Faculty of Engineering, Universiti Malaya, 50603 Kuala Lumpur, Malaysia\\}
\date{Accepted XXX. Received YYY; in original form ZZZ}

\pubyear{2025}

\begin{document}
	\label{firstpage}
	\pagerange{\pageref{firstpage}--\pageref{lastpage}}
	\maketitle
	
	\begin{abstract}
		In this work, {we develop, for the first time, a supervised classification framework with class-dependent rewards (CDR) to predict $\geq$M flares within 24 hr. We} construct multiple datasets, covering knowledge-informed features and line-of-sight (LOS) magnetograms. We {also} apply three deep learning models (CNN, CNN-BiLSTM, and Transformer) and three {CDR counterparts} ({CDR-CNN, CDR-CNN-BiLSTM, and CDR-Transformer}). {First, we} analyze the importance of LOS magnetic field parameters with the Transformer, then compare its performance using LOS-only, vector-only, and combined magnetic field parameters. {Second, we compare flare prediction performance based on CDR models versus deep learning counterparts. Third, we perform sensitivity analysis on reward engineering for CDR models. Fourth, we use the SHAP method for model interpretability.} Finally, we conduct performance comparison between our models and NASA/CCMC. The main findings are: (1) Among LOS feature combinations, R\_VALUE and AREA\_ACR consistently yield the best results. (2) Transformer achieves better performance with combined LOS and vector magnetic field data than with either alone. (3) Models using knowledge-informed features outperform those using magnetograms. (4) While CNN and CNN-BiLSTM outperform their {CDR} counterparts on magnetograms, {CDR-Transformer is slightly superior to} its deep learning counterpart when using knowledge-informed features. Among all models, {CDR-Transformer} achieves the best performance. {(5) The predictive performance of the CDR models is not overly sensitive to the reward choices. (6)}  Through SHAP analysis, {the CDR model tends to regard TOTUSJH as more important, while the Transformer tends to prioritize R\_VALUE more. (7)}  Under identical prediction time and active region (AR) number, the {CDR-Transformer} shows superior predictive capabilities compared to NASA/CCMC. 
	\end{abstract}
	
	
	
	
	\section{Introduction} \label{sec:intro}
	Solar flares are explosive phenomena in solar activity, and their energy release involves a sudden enhancement of high-temperature plasma, high-speed charged particle streams, and multi-band electromagnetic radiation. According to the classification criteria of the National Oceanic and Atmospheric Administration (NOAA) of the United States, solar flares are classified into five levels (A, B, C, M, and X) based on their peak X-ray flux, with each successive level representing a tenfold increase in energy. The high-energy particles and radiation released by high-intensity flares pose a serious threat to human activities, such as interfering with satellite navigation, causing disruptions to radio communications, damaging power transmission systems, and even endangering the safety of long-distance pipeline transportation \citep{boteler201921st,deshmukh2022decreasing}. Therefore, accurate and reliable prediction of solar flare is of great significance for ensuring the safety of human production activities and space exploration missions.
	
	In the past, many researchers mainly adopted statistical methods \citep{song2009statistical, mason2010testing,bloomfield2012toward,barnes2016comparison} and traditional machine learning techniques to predict the occurrence of solar flares. Among them, the well-known methods include artificial neural networks \citep{qahwaji2007automatic,ahmed2013solar,li2013solar,nishizuka2018deep}, K-nearest neighbor algorithm  \citep{huang2013improving,nishizuka2017solar}, Ensemble learning \citep{colak2009automated,huang2010short,guerra2015ensemble}, Support Vector Machine \citep{yuan2010automated,bobra2015solar,nishizuka2017solar,sadykov2017relationships}, Bayes network method \citep{yu2010short}, and random forest \citep{liu2017predicting,florios2018forecasting}. In recent years, image-based convolutional neural networks \citep[CNN; ][]{lecun2015deep}, as a learnable feature extractor, can automatically obtain features from image data without manual intervention and has become very popular in solar flare prediction. \citet{huang2018deep} and \citet{park2018application} constructed a CNN model for binary classification prediction of solar flares, respectively. \citet{zheng2019solar}, \citet{zheng2021hybrid} and \citet{zhang2024novel} proposed the hybrid convolutional neural network (H-CNN) model, the convolutional neural network (OVO-CNN) model, and the convolutional neural network (OAR-CNN) model for multi-class flare prediction within 24 hr, respectively. Similarly, Long Short-Term Memory (LSTM) model based on knowledge-informed features has been used in the research of flare prediction  \citep{chen2019identifying,liu2019predicting}. Another deep learning technique, the Transformer model \citep{vaswani2017attention}, employs a self-attention mechanism to capture global features more effectively. \citet{li2024prediction} constructed six solar flare prediction models using SHARP and high-energy-density dataset, demonstrating that the Transformer model significantly outperformed other approaches. Transformer-based models have great potential in the prediction of solar flares \citep{abduallah2023operational,ferreira2024transformers,pelkum2024forecasting,alshammari2024transformer}. {Although recent years have witnessed significant progress in solar flare prediction using supervised learning models such as CNN, LSTM, and Transformer, the flare prediction task has suffered from a severe class imbalance problem \citep{wan2021class} that may limit generalization ability on future data. \citet{mnih2015human} developed a deep Q Network (DQN) based on the deep reinforcement learning (DRL), which can directly learn successful policies from high-dimensional sensory inputs using end-to-end reinforcement learning. \citet{lin2020deep} used DRL to treat class imbalance problems. \citet{yi2023application} was inspired by the DRL and used LOS magnetograms to make daily "Yes" or "No" predictions for $\geq$M-class flares. However, the flare prediction task does not constitute a Markov decision process (MDP), so the complete DRL framework cannot be directly applied. We attempt to explore a method that draws inspiration from DRL but does not require constructing an MDP, thereby treating class imbalance and enabling high-precision flare prediction.}

	At present, the {knowledge-informed} features extracted from solar full-disk line-of-sight (LOS) magnetogram images have been widely used in flare prediction research (e.g., \citealt{boucheron2015prediction,boucheron2023solar}). \citet{al2015automated} and \citet{boucheron2023solar} calculated the magnetic field parameters from LOS photospheric magnetograms for solar flare prediction, respectively. \citet{sun2022predicting} used LOS magnetogram images and magnetic field parameters of active regions (ARs) to predict whether M-class or X-class flares would occur in ARs within 24 hr. Since the Solar Dynamics Observatory \citep[SDO; ][]{pesnell2012solar}
	began providing vector magnetogram images and vector magnetic field parameters in 2010, researchers have integrated LOS and vector magnetic field data as inputs to enhance solar flare prediction models. \citet{liu2019predicting} constructed the LSTM model and simultaneously took the LOS and vector magnetic field parameters as the model input, to forecast solar flares of $\geq$M5.0 class, $\geq$C-class, and $\geq$M-class within the next 24 hr. \citet{abduallah2023operational} also simultaneously took the LOS and vector magnetic field parameters as the input of the SolarFlareNet model to forecast solar flares of $\geq$M5.0 class, $\geq$C-class, and $\geq$M-class within the next 24 to 72 hr. \citet{pelkum2024forecasting} used the Transformer and simultaneously took LOS and vector magnetic field parameters as inputs to predict whether an AR would release a specific type of flare within the next 24 hr. Although more researchers use LOS and vector magnetic field data for flare prediction, no studies have explored the impact of three feature combinations, namely only LOS magnetic field parameters, only vector magnetic field parameters, and a combination of both, on model performance to date.

	\citet{sun2022predicting} comparatively studied the performance of flare prediction models based on LOS magnetogram images and magnetic field parameters, where the magnetic field parameters were exclusively derived from LOS magnetogram data. \citet{111} developed eight models for forecasting $\geq$M-class solar flares within 24 hours, conducting a comparative study of LOS magnetic field parameters and LOS magnetogram images. \citet{tang2021solar} and \citet{zheng2023multiclass} also comparatively studied the performance of flare prediction models based on LOS magnetogram images and magnetic field parameters. However, the parameters of the magnetic field are derived from both the LOS magnetogram images and the vector magnetogram images simultaneously. Since LOS magnetograms and vector magnetograms are two types of data of different qualities, directly comparing the performance of the model based on LOS magnetogram images with that of the model combining parameters from both LOS and vector magnetic fields may lead to an unfair evaluation of the LOS-based model performance. 
	
	{Inspired by the concept of DRL, we develop a supervised classification framework with class-dependent rewards (CDR) to treat class imbalance problem, and for the first time apply our newly CDR framework to the forecasting of $\geq$M-class flares within 24 hr. We} construct multiple datasets with magnetogram images and magnetic field parameters of ARs. We {also} apply three deep learning models including  CNN, CNN-BiLSTM, and Transformer models to make both categorical and probabilistic predictions for $\geq$M-class flares within 24 hr. Based on these three deep learning architectures, we  develop three {CDR} models, including {CDR-CNN, CDR-CNN-BiLSTM, and CDR-Transformer models} for flare prediction. {First,} we conduct an importance analysis of the LOS magnetic field parameters through Transformer. {Additionally,} we compare the categorical and probabilistic prediction performance of LOS magnetic field parameters, vector magnetic field parameters, and the combination of both using the Transformer model. { Second,} we compare the performance of the above flare prediction models based on magnetogram images (CNN, CNN- BiLSTM, {CDR-CNN, and CDR-CNN-BiLSTM}) and knowledge-informed features (Transformer and {CDR-Transformer}). {Third, we conduct a sensitivity analysis on reward engineering for the CDR models (CDR-Transformer, CDR-CNN, and CDR-CNN-BiLSTM).}  {Fourth,} we employ the SHapley Additive exPlanations \citep[SHAP;][]{NIPS} method to conduct interpretability studies on the deep learning model (Transformer) and the {CDR} model {(CDR-Transformer),} respectively. Finally, we compare the prediction performance of our models with that of the Community Coordination Modeling Center established by the National Aeronautics and Space Administration \citep[NASA/CCMC; ][]{hesse2001collisionless}.

	The remainder of this paper is organized as follows. Section \ref{sec:Data} describes the data, followed by the introduction of the method in Section \ref{sec:Meth}. The results are presented in Section \ref{sec:Res}, and the conclusions and discussions are provided in Section \ref{sec:conclu}.
	
	\section{Data}\label{sec:Data}
	
	Since April 30, 2010, the Helioseismic and Magnetic Imager \citep[HMI; ][]{schou2012design} has initiated regular observation missions and continuously provided high-resolution photospheric magnetograms from SDO \citep{pesnell2012solar}. \citet{bobra2014helioseismic} developed a derivative product, named Space Weather HMI Active Region Patch (SHARP), which automatically identifies and tracks ARs while generating corresponding LOS and vector magnetograms. In our study, we use a dataset that is not exactly the same as that of \citet{zheng2023multiclass}. This dataset includes 39 magnetic field parameters, comprising LOS and vector parameters, along with SHARP LOS magnetograms from May 1, 2010, to February 13, 2020. These 39 magnetic field parameters all belong to the knowledge-informed features. The relevant descriptions of knowledge-informed features are shown in Tables \ref{tab1} and \ref{tab2}. This dataset provides 31 LOS magnetic field parameters and 8 vector magnetic field parameters, while \citet{zheng2023multiclass} only contained 10 magnetic field parameters. The SHARP LOS magnetogram data provided by this dataset is consistent with that of \citet{zheng2023multiclass}. This dataset also includes labeled flare information (i.e., No-flare/C/M/X) based on the solar flare events catalog provided by the Geostationary Operational Environmental Satellite (GOES) and the Solar Geophysical Data solar event reports. The steps for collecting four-class SHARP data in the dataset are the same as those used by \citet{li2020predicting} and \citet{zheng2023multiclass}. The magnetogram images and magnetic field parameters are collected every 36 minutes to ensure sufficient variation between the nearest AR samples.

	This dataset contains SHARP magnetograms and magnetic field parameters of 808 ARs. These ARs are classified as 398 No-flare (weaker than C1.0) class, 299 C-class, 94 M-class, and 17 X-class, with each AR represented by 40 samples. This dataset includes a large number of ARs, spanning the entire 24th solar activity cycle. Both the magnetograms and the magnetic field parameters are time series, and their time and AR numbers are consistent. We shuffle the NOAA AR numbers of solar flares categorized as No-flare/C/M/X, then split the data to generate training, validation, and testing sets. This data-splitting process is repeated 10 times using different random seeds to create 10 separate datasets. This method ensures that in a single dataset, there is neither sample overlap nor AR overlap among the training, validation, and testing sets. In some cases, the recognition and extraction algorithms would identify the region composed of multiple NOAA ARs as a coherent magnetic structure (i.e., a HARP). For the sake of simplicity, such HARPs are excluded from the samples \citep{bobra2014helioseismic}. \citet{wang2020predicting} also indicated that multiple ARs exist in $20\%$ of SHARP. If the maximum flare category is assigned to time series of flares that may be omitted or misattributed to SHARP, this could introduce potential errors. Unlike the datasets containing both single and multiple ARs \citep{zheng2023multiclass}, we screen the validation and testing sets from the 10 separate datasets, retaining only single AR samples. To ensure sufficient training data for the model, the training set was not screened. Eventually, we create 10 cross-validation (CV) datasets in our work, which are used for model training, validation, and testing in Sections \ref{subsec:significance} and \ref{subsec:per comprison}. Table \ref{tab3} provides the distribution of the number of ARs from the 10 CV datasets used in this study, including magnetogram images and knowledge-informed features. The number of samples for each category in the training, validation, and testing sets is 40 times the corresponding number of ARs. The “No-flare/C/M/X” in Table \ref{tab3} represents the number of ARs corresponding to their respective categories among the training, validation, and testing sets. It is worth noting that the 10 training sets in Table \ref{tab3} only have the same quantity distribution, but the samples of the ARs in each training set are not exactly the same. Furthermore, Table \ref{tab1} lists 31 knowledge-informed features calculated from SHARP LOS magnetograms used in this study. Among the 31 LOS knowledge-informed features we calculated, the  R\textunderscore VALUE and AREA\textunderscore ACR features derive from  \citet{schrijver2007characteristic} and \citet{bobra2015solar}, with the remaining 29 features sourced from \citet{al2015automated} and \citet{boucheron2023solar}. \citet{111} also conducted a related study on these 31 knowledge-informed features. The 8 vector knowledge-informed features used in this study are derived from SHARP vector magnetograms and listed in Table \ref{tab2}. Further detailed explanations of these features are provided by \citet{liu2019predicting} and \citet{abduallah2023operational}. Considering the differences in the scale and unit of the magnetic field features, we adopt a standardized method to process these features before inputting them into the flare prediction model.

	\section{Method} \label{sec:Meth}
	{We propose a CDR framework for flare prediction for the first time.} We {also} develop three deep learning models (CNN, CNN-BiLSTM, and Transformer) and three {CDR} models {(CDR-CNN, CDR-BiLSTM, and CDR-Transformer)} for predicting  $\geq$M-class solar flares within 24 hr, and comparatively study the prediction performance of these six models. The algorithmic model structures of deep-learning models and {CDR models} are identical. The difference lies in the training methods.
	
	\subsection{ {A class-dependent reward framework}} \label{subsec:DRL}
	{DQN is a representative algorithm in DRL that integrates Q-learning with deep neural networks \citep{mnih2015human}. \citet{lin2020deep} further demonstrated through experiments that a carefully designed reward scheme enables DRL to effectively treat the challenge of extreme class imbalance. Inspired by the reward scheme in DRL, this study initially considered introducing a complete DRL framework to treat the class imbalance of the flare prediction task, thereby achieving high-precision flare prediction. However, since the flare prediction task does not constitute an MDP, a complete DRL framework cannot be directly employed. Therefore, we introduce reward schemes into solar flare prediction and propose the CDR framework for the first time. The CDR framework guides the model to more accurately capture discriminative features across different categories by assigning distinct immediate rewards based on prediction outcomes during training process. }
	
\begin{figure*}
		\centering
		\includegraphics[width=0.96\textwidth]{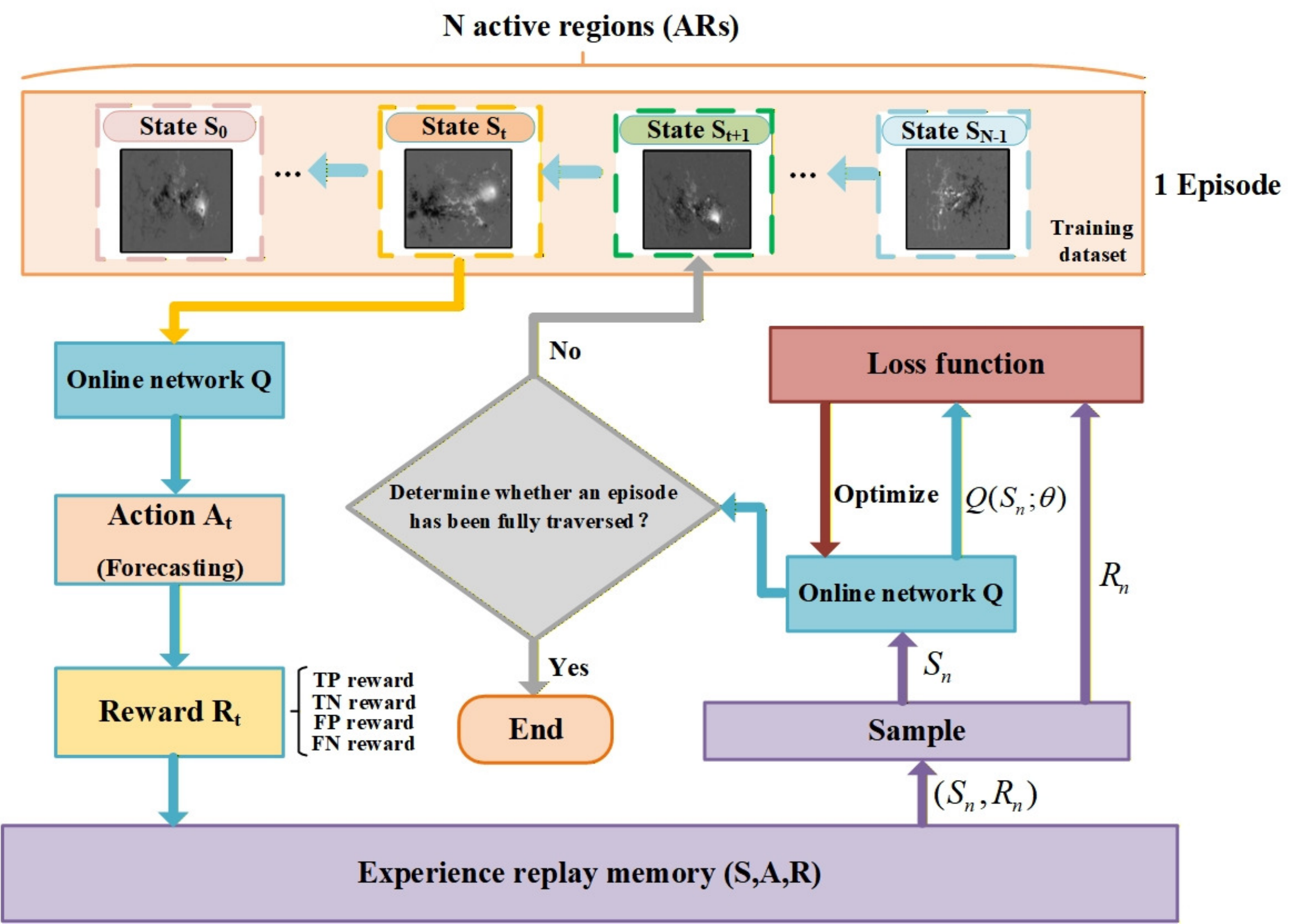}
		\caption{{Training process of the CDR framework.}}\label{figure1}
	\end{figure*}

	{Figure \ref{figure1}  shows the training process of the CDR framework. We define an AR as a state (denoted as $S$), where each AR consists of time series samples over a consecutive 24-hour period, and each sample is composed of 10 knowledge-informed features or magnetograms. We use the predicted value of whether the AR will flare as the action (denoted as $A$). Specifically, when the AR is predicted to flare, $A$=1; otherwise, $A$=0. The immediate reward value (denoted as $R$) in the reward scheme includes the True Positive (TP) reward, True Negative (TN) reward, False Positive (FP) reward, or False Negative (FN) reward. When the true label of an AR is positive and the model correctly predicts it as positive, a TP reward is given. Similarly, a TN reward is given when the true label of an AR is negative and the model correctly predicts it as negative. If the true label of an AR is negative but the model incorrectly predicts it as positive, an FP reward is assigned. Conversely, an FN reward is given when the true label of an AR is positive but the model predicts it as negative. The CDR framework can set up multiple episodes for training, with each episode fully traversing all AR data in the training set. In addition, we introduce the experience replay memory from the DRL into the CDR framework to store state, action, and immediate reward ($S$, $A$, $R$). This allows the online network Q to perform random sampling of ($S$, $A$, $R$) from the experience replay memory and feed them into the online network Q for training, thereby improving the stability of model training and the sample utilization rate.}
		
	{As shown in Figure \ref{figure1}, the online network Q has the same model structure as CNN, CNN-BiLSTM, or Transformer in Section \ref{subsec:flare pre}. First, the state $S_t$ is input into the online network Q, which outputs the predicted probability value {$Q(S_t, \theta)$}, thereby predicting whether a flare will occur in the AR. Based on the prediction result, an immediate reward is obtained. Then, the action and immediate reward value corresponding to the current AR $Q(S_t, A_t, R_t)$ are stored in the experience replay memory. When the number of ARs in the experience replay memory exceeds the given batch size, the online network Q randomly selects $(S_n, R_n)$ ($n$=1, 2, …, batch size) from the experience replay memory. The $S_n$ is then input into the online network Q, which outputs {$Q(S_n, \theta)$}. Based on the {Formula} (\ref{eq:4})  in Section \ref{subsec:flare pre}, the loss function value is calculated, and the parameters of the online network Q are updated. After the parameter update is complete, the online network Q checks whether a full episode has been traversed. If not, we input the next state $S_{t+1}$ into the online network Q and repeat the above process for training. If a full episode has been traversed, the current episode ends, and the training for the next episode begins, continuing until all episodes are completed.}

	\subsection{Flare prediction model} \label{subsec:flare pre}
	
\begin{figure*}
		\centering
		\includegraphics[width=\textwidth]{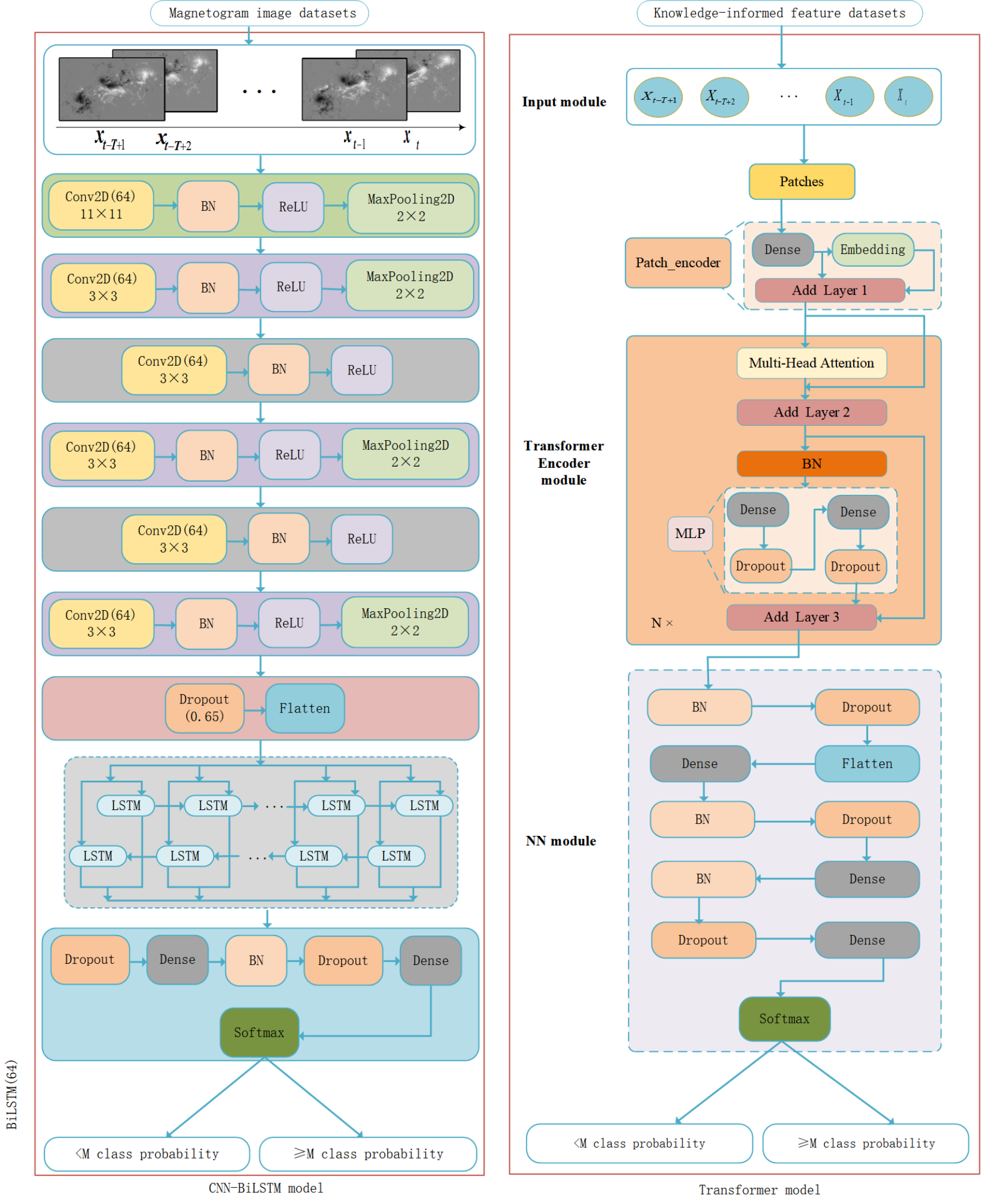}
		\caption{The architecture of flare prediction models.}\label{figure2}
	\end{figure*}

	In this study, we apply the CNN and CNN-BiLSTM models consistent with \citet{111}, develop a Transformer flare forecasting model, and comparatively study the forecasting performance of these three models. Figure \ref{figure2} shows the model architectures of CNN-BiLSTM and Transformer. The CNN-BiLSTM mainly consists of a CNN and a bidirectional LSTM network. CNN contains six convolutional layers. Each convolutional layer consists of an 11×11 or 3×3 convolutional kernel, batch normalization \citep[BN; ][]{pmlr-v37-ioffe15}, and a modified linear unit \citep[ReLU; ][]{nair2010rectified}. In all the convolutional layers, four of them also include a 2×2 MaxPooling layer. Next, after passing through the Dropout function and the Flatten layer, the data is input into the BiLSTM network. The BiLSTM network consists of a forward LSTM layer and a reverse LSTM layer. At time $t$, the output of the forward LSTM layer is represented as $\overrightarrow{h_t}$, while the output of the reverse LSTM layer is represented as $\overleftarrow{h_t}$. The output representations of the above two LSTM layers are as follows:
	
	\begin{equation}
		h_t = \left[ \overrightarrow{h_t}, \overleftarrow{h_t} \right].
		\label{eq:1}
	\end{equation}
	Subsequently, the data flows through a Dropout function, a Dense layer, a BN layer, another Dropout function, and a second Dense layer sequentially. Finally, it is passed to the Softmax activation function, thereby generating the prediction results of the CNN-BiLSTM. Furthermore, the architecture of the CNN model is derived by removing the BiLSTM network and its subsequent Dropout function based on the structure of the CNN-BiLSTM.	
	
	The Transformer consists of the Input module, the Patches module, the Patch\_encoder module, the Transformer encoder module, and the Neural Network (NN) modules, which is shown in Figure \ref{figure2}. The input sequence of the model is usually represented as $[X_{t-T+1}, X_{t-T+2}, \ldots, X_{t-1}, X_t]$
	, where $X_{t}$ represents the input value captured at time  $t$, $X_{t-1}$
	represents the input value captured at time $t-1$, and $T$ represents the time step. The data output by the Input module is converted into dimensions suitable for processing by the Transformer through the Patches module and the Patch\textunderscore encoder module. The Patch\textunderscore encoder module integrates a Dense layer with an Embedding layer and an Add Layer 1. The Add Layer 1 sums the input to the Embedding layer and its output, forming a residual connection. Each Transformer encoder module consists of one Multi-Head Attention (MHA) layer, two Add Layers, one BN layer, and one Multi-Layer Perceptron (MLP) layer. MHA layer utilizes multiple independent self-attention mechanisms to simultaneously focus on different parts of the input sequence and capture diverse dependencies \citep{DoubleDQN}. Add Layer 2 performs a residual connection by summing the input and output of the MHA layer, thereby promoting better model training and optimization \citep{he2016deep}. BN layer enhances training stability and convergence speed while providing a regularization effect that improves model generalization \citep{zerveas2021transformer}. The MLP layer further transforms and extracts features to represent them more comprehensively, which comprises two Dense layers and two Dropout functions. Add Layer 3 additionally implements a residual connection by summing the input of the BN layer with the output of the MLP layer, then passing the combined result to the subsequent Transformer encoder module. In our work, we have stacked a total of 4 Transformer encoder modules. The NN module consists of three BN layers, three Dense layers, one Flatten layer, three Dropout functions, and one Softmax activation function. Ultimately, the Transformer model classifies the samples into $<$M-class or $\geq$M-class based on the magnitude of the output probability.
	
	Based on the above three deep learning model architectures, we also develop three {CDR} models, including {CDR-CNN, CDR-CNN-BiLSTM, and CDR-Transformer}, to explore the influence of {CDR} on the performance of solar flare prediction. Tables \ref{tab4} and \ref{tab5} present the parameter configurations of deep learning and {CDR} models, respectively. The inputs of the CNN and {CDR-CNN} are point-in-time SHARP image data; the inputs of the CNN-BiLSTM and {CDR-CNN-BiLSTM } are time-series SHARP image data; and the inputs of the Transformer and {CDR-Transformer} are time-series knowledge-informed features. We implement the network using the Keras framework with the TensorFlow backend \citep{199317}. We trained the model using either Stochastic Gradient Descent \citep[SGD; ][]{lecun2015deep} or adaptive moment estimation \citep[Adam; ][]{kingma2014adam} to minimize the cross-entropy loss function. For the deep learning model, we adopt the weighted cross-entropy function as the loss function, as shown below:

	\begin{equation}
		L_{\text{DL}} = \sum_{n=1}^{{B}} \sum_{k=1}^{K} w_k y_{nk} \log_{e}(\hat{y}_{nk}),
		\label{eq:2}
	\end{equation}

	\begin{equation}
		\omega_k = \frac{N_{\text{total}}}{K \cdot N_k},
		\label{eq:3}
	\end{equation}
	let $K$=2 be the number of categories, and ${B}$ indicates the {batch size} size during training. Furthermore, $y_{nk}$ and $\hat{y}_{nk}$ represent the target label and predicted output of the $k$th
	class in the forward propagation process, respectively. $\omega_k$ represents the weight of the $k$th category used for the weighted loss function. $N_{total}$ represents the total number of training samples, while $N_{k}$  represents the number of training samples in the $k$th category. In the {CDR} model, the loss function is defined as follows:
	
	\begin{equation}
	{L_{\text{CDR}}(\theta) = \sum_{n=1}^{B} y_n \log_e \left( Q(S_n; \theta) \right),}
		\label{eq:4}
	\end{equation}

	\begin{equation}
	{y_n = R_n = }
	\begin{cases}
		{\text{TP reward, } A_n = 1 \text{ and } S_n \in L_P} \\
		{\text{TN reward, } A_n = 0 \text{ and } S_n \in L_N} \\
		{\text{FP reward, } A_n = 1 \text{ and } S_n \in L_N} \\
		{\text{FN reward, } A_n = 0 \text{ and } S_n \in L_P}
	\end{cases}{,}
		\label{eq:5}
	\end{equation}
	{where $\theta$ represents the parameters of the current network. We define the $n$th AR as a state $S_n$. $Q(S_n, \theta)$ is the predicted probability value, and $y_n$ is equal to the immediate reward value $R_n$ for the $n$th AR. $A_n$ is the predicted value indicating whether a flare occurs in the $n$th AR. $B$ is the batch size, consistent with that in Formula (\ref{eq:2}). $L_P$ represents the true label of the AR belonging to the positive class, while $L_N$ represents the true label of the AR belonging to the negative class.}

	\section{Results} \label{sec:Res}
	
	To fairly evaluate the model from different perspectives, we consider both its categorical and probabilistic performance for $\geq$M-class flare prediction. {We evaluate categorical forecasting performance using the True Skill Statistic \citep[TSS; ][]{bib41}  and assess probabilistic prediction performance with the Brier Skill{Score}  \citep[BSS;][]{wilks2011statistical}.} 
	
	{We} describe the prediction results using a confusion matrix to evaluate the categorical forecasting performance of the models. The confusion matrix consists of TP, TN, FP, and FN, which are essential for calculating metrics such as recall, FPR, and TSS. These metrics are summarized below:

	\begin{equation}
		\text{Recall} = \frac{\text{TP}}{\text{TP} + \text{FN}},
		\label{eq:7}
	\end{equation}

	\begin{equation}
		\text{FPR} = \frac{\text{FP}}{\text{FP} + \text{TN}},
		\label{eq:8}
	\end{equation}
	
	\begin{equation}
		\text{TSS} = \text{Recall} - \text{FPR}.
		\label{eq:9}
	\end{equation}
	Recall assesses the model's predictive ability to identify positive class samples, while FPR quantifies its error rate in negative class predictions. The score range of Recall is from 0 to 1, where 1 indicates the best performance. The FPR score range is also from 0 to 1, with 0 representing the ideal outcome. The TSS ranges from -1 to 1, where 1 denotes perfect skill, 0 indicates no discriminatory ability, and -1 represents the worst possible {performance. In} the flare prediction task, due to the high imbalance of the samples, the TSS score which is insensitive to category imbalance, is adopted as the primary evaluation criterion \citep{bloomfield2012toward,bobra2015solar}. Therefore, we use TSS as the primary evaluation metric for categorical prediction performance, with the other metrics serving as secondary indicators. In the categorical prediction for all sections except Section \ref{subsec:NASA}, the probability threshold is the default value of 0.5. 
	
	{We also} use BSS to evaluate the probabilistic forecasting performance of the model. BSS is calculated based on the Brier Score \citep[BS; ][]{brier1950verification} as follows:
	\begin{equation}
		BS_k = \frac{1}{N} \sum_{n=1}^{N} (y_{nk} - \hat{y}_{nk})^2,
		\label{eq:10}
	\end{equation}
	
	\begin{equation}
		BSS_k = 1 - \frac{BS_k}{\frac{1}{N} \sum_{n=1}^{N} (y_{nk} - \bar{y}_k)^2}.
		\label{eq:11}
	\end{equation}For the testing samples, $y_{nk}$ and $\hat{y}_{nk}$ represent the target label and the predicted probability of the $k$th class, respectively. The target variable $y_{nk}$ equals 0 if no flare occurs and equals 1 if a flare occurs. $N$ is the total number of testing samples. $\bar{y}_k$ represents the average value of the target label for the $k$th class in the testing samples, which corresponds to the average occurrence frequency of the $k$th class. The value range of BS is 0 to 1, where a result closer to 0 indicates better probabilistic prediction performance of the model. The BSS score ranges from $-\infty$ to 1, with 1 indicating the best probabilistic forecast.

	\subsection{Knowledge-informed Feature Significance Analysis} \label{subsec:significance}
	
	In our work, during the process of model training for the categorical prediction, the TSS score is monitored on the validation set after each epoch, and the model is saved when the TSS score reaches its maximum. Similarly, during the process of model training for the probabilistic prediction, the BSS score on the validation set is recorded after each epoch, and the model is saved when the maximum BSS score is reached. In categorical and probabilistic prediction, we respectively use the saved models for testing on the testing set. Finally, we repeat this process on 10 separate training, validation, and testing sets from 10 CV datasets, ultimately obtaining the results of categorical and probabilistic prediction, and providing the means and standard deviations of these results.

	We analyze the importance of the combination of knowledge-informed features using the TSS and BSS, respectively. Based on the possible combination of feature categories, the 31 knowledge-informed features calculated from the LOS magnetograms are divided into 10 groups. We use the dataset comprising these 10 groups of features to train, validate, and test the Transformer model. Through evaluating its categorical and probabilistic prediction performance, we thereby determine the importance of the 10 groups of features. Table \ref{tab6} presents the TSS score for categorical prediction of $\geq$M-class flares using the Transformer model, as well as the BSS score for probabilistic prediction, involving a total of 10 groups of LOS knowledge-informed features. In the categorical prediction shown in Table \ref{tab6}, the TSS score of the model on the feature combination of  R\textunderscore VALUE and AREA\textunderscore ACR
	is 0.697±0.137, which is higher than the TSS score of the model on other feature combinations. Similarly, in probabilistic prediction shown in Table \ref{tab6}, the BSS score of the model on the feature combination of  R\textunderscore VALUE and AREA\textunderscore ACR
	is 0.448±0.079, which is also superior to the BSS score on other feature combinations. Overall, our results indicate that among the 10 groups of features containing 31 LOS knowledge-informed features, our Transformer model exhibits the optimal performance in predicting $\geq$M-class flares when using the feature combination of  R\textunderscore VALUE and AREA\textunderscore ACR
	. The large flares frequently occur near strong and highly sheared the polarity inversion line \citep[PIL; ][]{vasantharaju2018statistical,dhakal2023causes}, and the  R\textunderscore VALUE 
	focuses on the regions near the PIL. Furthermore, the significance of  R\textunderscore VALUE
	in the solar flare predictions by other scholars has also been confirmed (e.g., \citealt{liu2017predicting,li2024prediction, xiang2024analysis, 111}). Therefore, in the following section, the model incorporating LOS knowledge-informed features, namely the Transformer, exclusively utilizes the feature combination of  R\textunderscore VALUE and AREA\textunderscore ACR
	.

	In addition to the LOS magnetic field parameters, the vector magnetic field parameters are also important knowledge-informed features. In our work, we independently input three distinct combinations of magnetic field parameters into the Transformer model: 2 optimal LOS magnetic field parameters (i.e.,  R\textunderscore VALUE and AREA\textunderscore ACR), 8 vector magnetic field parameters, and the combination of both. Each combination is separately used for training, validation, and testing. Table \ref{tab7} shows the TSS and BSS scores for $\geq$M-class flare prediction using the Transformer on the testing set with three distinct combinations of magnetic field parameters. In the categorical prediction shown in Table \ref{tab7}, the TSS scores of 2 LOS magnetic field parameters, 8 vector magnetic field parameters, and the combination of both are 0.697±0.137, 0.719±0.090, and 0.814±0.029, respectively. In probabilistic prediction, the BSS scores of 2 LOS magnetic field parameters, 8 vector magnetic field parameters, and the combination of both are 0.448±0.079, 0.403±0.098, and 0.452±0.095, respectively. The model using the combination of LOS and vector magnetic field parameters yields higher TSS and BSS scores than models when using either type alone. Our results indicate that among the magnetic field parameters, the combination of LOS and vector magnetic field parameters are the most effective feature combination for $\geq$M-class flare prediction. The LOS magnetic field parameters provide key information about the strength and variation of the longitudinal magnetic field. For example, the  R\textunderscore VALUE 
	\citep{schrijver2007characteristic,li2024prediction} can be used to monitor rapid changes in magnetic flux. The vector magnetic field parameters offer information about the transverse magnetic field, such as magnetic shear angle, current density, and magnetic helicity \citep{leka2003photospheric,fisher2013channel,zheng2023multiclass}. Combining the LOS and vector magnetic field parameters could describe the magnetic field state of the solar AR more comprehensively. Furthermore, variations in the magnetic field are closely associated with solar flares. Therefore, in this study, we integrate LOS and vector magnetic field parameters to significantly enhance the flare prediction performance of the model.

	\subsection{Performance comparison of prediction models} \label{subsec:per comprison}
	
	Since the LOS and vector magnetic field parameters are derived from data with differing quality levels, directly comparing a model based on both LOS and vector parameters with one based on LOS magnetogram images may introduce bias. To ensure a fair evaluation, we directly compare the prediction performance of the model based on 2 LOS magnetic field parameters with that of the model using LOS magnetogram images. In Section \ref{subsec:per comprison}, the model utilizing the optimal combination of LOS and vector magnetic field parameters is designated as Transformer-10, while the model employing 2 LOS magnetic field parameters is designated as Transformer-2.
	
	The flare prediction models based on image data are trained, validated, and tested on the magnetogram images from 10 CV datasets. Similarly, the models based on knowledge-informed features are trained, validated, and tested on the magnetic field parameters from 10 CV datasets. Table \ref{tab8} presents the categorical and probabilistic prediction performance of deep learning and {CDR} models across 10 testing sets from the 10 CV datasets. In deep learning, the CNN and CNN-BiLSTM models, which are based on image data, are consistent with those presented by \Citet{111}. The results from these models, also consistent with \Citet{111}, achieve TSS scores of 0.543±0.101 and 0.682±0.085, respectively, and BSS scores of 0.242±0.100 and 0.443±0.105, respectively. For the models utilizing knowledge-informed features, the Transformer-10 and Transformer-2 models, achieve TSS scores of 0.814 ± 0.029 and 0.697 ± 0.137, with corresponding BSS scores of 0.452 ± 0.090 and 0.448 ± 0.079, respectively. We first perform a direct comparison between the models based on LOS magnetogram images and those based on LOS magnetic field parameters. The Transformer-2, trained on 2 LOS magnetic field parameters, outperforms the CNN and CNN-BiLSTM trained on the LOS magnetogram images, achieving significantly higher TSS and BSS scores. Moreover, the TSS and BSS scores of the deep learning model (Transformer-10), which utilizes the combination of LOS and vector magnetic field parameters, surpass those of the deep learning model (Transformer-2) based solely on LOS magnetic field parameters. Consequently, deep learning models using LOS magnetic field parameters and the combination of LOS and vector magnetic field parameters (i.e., Transformer-2 and Transformer-10) achieve higher TSS and BSS scores compared to image-based deep learning models (i.e., CNN and CNN-BiLSTM). Among these models, the deep learning model using the combination of LOS and vector magnetic field parameters (Transformer-10) demonstrates the best prediction performance. Therefore, deep learning models utilizing knowledge-informed features outperform those based on magnetogram images in both categorical and probabilistic predictions.
	
	In {CDR models}, for the {CDR-CNN} and {CDR-CNN-BiLSTM} models based on image data, the TSS scores are {0.495±0.153} and {0.453±0.132} respectively, and the BSS scores are {0.076±0.073} and {0.017±0.122} respectively. For the {CDR-Transformer-10} and {CDR-Transformer-2} models based on knowledge-informed features, the TSS scores are {0.829 ±0.059} and {0.714±0.062} respectively, and the BSS scores are {0.489±0.120} and {0.470±0.070} respectively. In categorical and probabilistic prediction, the TSS and BSS scores of the {CDR} model based on the 2 LOS magnetic field parameters ({CDR-Transformer-2}) are both higher than those of the {CDR} models based on magnetogram images ({CDR-CNN }and {CDR-CNN-BiLSTM}). Furthermore, the TSS and BSS scores of the {CDR} model using the combination of LOS and vector magnetic field parameters ({CDR-Transformer-10}) are both higher than those of the {CDR} model using LOS magnetic field parameters ({CDR-Transformer-2}). {CDR-Transformer-10}, which uses the combination of LOS and vector magnetic field parameters, achieves the best prediction performance among all {CDR} models. Therefore, {CDR} models based on knowledge-informed features are superior to those based on magnetogram images.
	
	In addition to the influence of data type (i.e., image data and knowledge-informed features) on model performance, the incorporation of the {CDR framework} also affects model performance. As presented in Table \ref{tab8}, the deep learning models CNN and CNN-BiLSTM, which utilize magnetogram images, achieve TSS scores of 0.543 ± 0.101 and 0.682 ± 0.085, respectively, and BSS scores of 0.242 ± 0.140 and 0.443 ± 0.105, respectively. In contrast, the {CDR} models {, CDR-CNN} and {CDR-CNN-BiLSTM}, also based on magnetogram images, achieve TSS scores of {0.495 ± 0.153}and {0.453 ± 0.132}, respectively, and BSS scores of {0.076 ± 0.073} and {0.017 ± 0.122}, respectively. These results indicate that deep learning models based on magnetogram images (CNN and CNN-BiLSTM) achieve higher TSS and BSS scores compared to {CDR} counterparts based on magnetogram images ({CDR-CNN} and {CDR-CNN-BiLSTM}). Consequently, in both categorical and probabilistic predictions, deep learning models leveraging image data outperform their {CDR} counterparts. However, the deep learning models Transformer-10 and Transformer-2, which utilize knowledge-informed features, achieve TSS scores of 0.814 ± 0.029 and 0.697 ± 0.137, respectively, and BSS scores of 0.452 ± 0.090 and 0.448 ± 0.079, respectively. In contrast, the {CDR} models, {CDR-Transformer-10} and {CDR-Transformer-2}, also based on knowledge-informed features, achieve TSS scores of { 0.829 ± 0.059} and {0.714 ± 0.062}, respectively, and BSS scores of {0.489 ± 0.120} and {0.470 ± 0.070}, respectively. { We conduct a series of paired t-tests to evaluate whether the CDR models based on knowledge-informed features (CDR-Transformer-10 and CDR-Transformer-2) significantly outperform the deep learning models (Transformer-10 and Transformer-2) in prediction performance. The paired t-test results between the CDR-Transformer and Transformer models are shown in Table \ref{tab9}. As shown in Table \ref{tab9}, the CDR-Transformer-10 model, compared with the Transformer-10, yields p-value of 0.262 for TSS and 0.094 for BSS, neither of which reaches statistical significance (p-value < 0.05). Similarly, the CDR-Transformer-2 model, compared to the Transformer-2 model, yields p-value of 0.366 for TSS and 0.431 for BSS, neither of which reaches the statistical significance (p-value < 0.05).} Consequently, the {CDR} models based on knowledge-informed features ({CDR-Transformer-10} and {CDR-Transformer-2}) { slightly outperform} their deep learning counterparts (Transformer-10 and Transformer-2) {in terms of TSS and BSS scores. Although the CDR models (CDR-Transformer-10 and CDR-Transformer-2) are slightly superior to the deep learning models (Transformer-10 and Transformer-2) in categorical and probabilistic prediction performance, they effectively guide the learning process by assigning different immediate reward values, enabling better capture of key features from samples of different categories. Moreover, when treating the class imbalance problem in solar flare forecasting, the CDR models demonstrate unique advantages.}
	
	The optimal {CDR-Transformer-10} model from our study is compared with the optimal iTransformer model from \citet{111}, both of which use knowledge-informed features with identical AR number and distribution. {The iTransformer \citep{liu2023} is an inverted Transformer architecture specifically designed for multivariate time-series forecasting. Unlike the standard Transformer architecture adopted in this study, the iTransformer treats individual variates as token and performs attention calculation along the variate dimension, thereby capturing inter-variate correlations efficiently. In contrast, the standard Transformer architecture proposed in this study follows the approach by treating the ten knowledge-informed features at each time step as token and applying attention calculation along the temporal dimension. This effectively captures temporal dependencies among time series samples within the AR.}
	
	{As} shown in Table \ref{tab8}, the iTransformer model achieves TSS and BSS scores of 0.768±0.072 and 0.513±0.063, respectively, while the {CDR-Transformer-10} model achieves TSS and BSS scores of {0.829±0.059} and {0.489±0.120}, respectively. Therefore, the optimal {CDR-Transformer-10} model in this study outperforms the iTransformer model in TSS score, while its BSS score is comparable to that of iTransformer. Thus, in terms of categorical and probabilistic forecasting, the performance of the { CDR-Transformer-10} model is superior to or comparable to that of the optimal iTransformer model of \citet{111}. Although { CDR-Transformer-10} and iTransformer use datasets with identical AR number and distribution, { CDR-Transformer-10} outperforms iTransformer in categorical forecasting tasks. This superior performance, beyond differences in training methods and model architectures, stems from its richer feature set, which enhances its predictive capabilities. The iTransformer model only utilized the R\_VALUE feature, whereas { CDR-Transformer-10} employs a combination of 10 LOS and vector features. These vector features are crucial as they could capture magnetic field topological structures, which are closely related to flare occurrences. This allows { CDR-Transformer-10} to better characterize the physical processes associated with flares, thereby improving its categorical forecasting performance.

	In summary, for both categorical and probabilistic predictions, the performance of deep learning models (Transformer-10 and Transformer-2) and {CDR} models ({CDR-Transformer-10} and {CDR-Transformer-2}) based on knowledge-informed features surpasses that of deep learning models (CNN and CNN-BiLSTM) and {CDR} models ({CDR-CNN }  and {CDR-CNN-BiLSTM}) based on magnetogram images. The knowledge-informed features, such as  R\textunderscore VALUE, represent physical parameters that are strongly correlated with flare occurrence, enabling models to more effectively exploit established physical relationships between solar activity indicators and flare events. In contrast, image-based models depend on automatic feature extraction from magnetogram images, which typically lacks a physical basis and may limit their capacity to identify critical features. As a result, the models based on knowledge-informed features generally achieve superior performance compared to those based on magnetogram images. In addition, the prediction performance of deep learning models based on magnetogram images (CNN and CNN-BiLSTM) surpasses that of {CDR} counterparts ({CDR-CNN} and {CDR-CNN-BiLSTM}){, while CDR} models utilizing knowledge-informed features ({CDR-Transformer-10} and {CDR-Transformer-2}) {slightly } outperform their deep learning counterparts (Transformer-10 and Transformer-2). This disparity may arise because deep learning excels at representation learning and feature extraction in high-dimensional image data, while {CDR} more effectively leverages domain knowledge through reward mechanisms and strategy optimization in low-dimensional  structured physical feature spaces.
	
	\subsection{{Sensitivity Analysis on Reward Engineering}} \label{subsec:sen}
	{To verify whether the superiority of the optimal CDR-Transformer-10 model is overly sensitive to immediate reward choices, we apply numerical perturbations to the TP, TN, FP, and FN rewards. We then retrain, validate, and test the model on the 10 CV datasets and compare the TSS and BSS scores of the perturbed model with those of the unperturbed model.}
	
	{As shown in Table \ref{tab5}, the optimal CDR-Transformer-10 model continuously optimizes the immediate reward values through experiments, ultimately selecting a set of parameters with the best classification prediction performance as the benchmark for rewards. Specifically, the TP reward is equal to 10, the TN reward is equal to 4, the FP reward is equal to -20, and the FN reward is equal to -15. As shown in Table \ref{tab8}, the TSS score of the optimal CDR-Transformer-10 model is 0.829 ± 0.059, and its BSS score is 0.489 ± 0.120. First of all, only the TP reward is varied, while the TN, FP, and FN rewards are kept constant. Table \ref{tab10} illustrates the effects of applying a perturbation (step size=1) to the TP reward on the TSS and BSS scores of the CDR-Transformer-10 model, with the benchmark indicated in bold font. We use the optimal TP reward of 10 identified by the CDR-Transformer-10 model as the benchmark value and decrease it sequentially in step size of 1. The corresponding TSS and BSS scores are shown in Table \ref{tab10}. The TSS scores of the model are 0.801 ± 0.082, 0.785 ± 0.076, 0.800 ± 0.064, 0.811 ± 0.062, and 0.803 ± 0.048, while the BSS scores are 0.482 ± 0.136, 0.481 ± 0.136, 0.475 ± 0.148, 0.467 ± 0.160, and 0.461 ± 0.172, respectively. When the TP reward is increased by step size of 1 at a time relative to the benchmark reward value, the TSS score of the model is 0.811 ± 0.077, 0.800 ± 0.067, 0.784 ± 0.079, 0.754 ± 0.096, and 0.793 ± 0.082. The BSS scores are 0.487 ± 0.107, 0.490 ± 0.096, 0.496 ± 0.092, 0.499 ± 0.091, and 0.492 ± 0.119.}
	
	{The TN reward is only changed, while the TP, FP, and FN rewards remain unchanged. Table \ref{tab11} shows the effects of applying a perturbation (step size=1) to the TN reward on the TSS and BSS scores of the CDR-Transformer-10 model. As shown in Table \ref{tab11}, we successively decrease the TN reward from its benchmark in step size of 1. The corresponding TSS scores are 0.815 ± 0.062, 0.826 ± 0.061, and 0.797 ± 0.067, respectively, while the BSS scores are 0.489 ± 0.119, 0.492 ± 0.117, and 0.492 ± 0.119. When the TN reward is successively increased by a step size of 1 relative to the benchmark reward value, the TSS scores are 0.826 ± 0.063, 0.821 ± 0.070, and 0.821 ± 0.070, while the BSS scores are 0.488 ± 0.121, 0.486 ± 0.123, and 0.488 ± 0.116.}
	
	{Furthermore, only the FP reward is changed, whereas the TP, TN, and FN rewards remain unchanged. Table \ref{tab12} shows the effects of applying a perturbation (step size=1) to the FP reward on the TSS and BSS scores of the CDR-Transformer-10 model. As shown in Table \ref{tab12}, we successively decrease the FP reward from its benchmark in step size of 1. The TSS scores are 0.824 ± 0.057, 0.808 ± 0.058, 0.810 ± 0.061, 0.809 ± 0.066, and 0.816 ± 0.069. The BSS scores are 0.487 ± 0.124, 0.484 ± 0.126, 0.486 ± 0.122, 0.483 ± 0.126, and 0.481 ± 0.128. When the FP reward increases by step size of 1 at a time compared to the benchmark reward value, the TSS scores are 0.819 ± 0.072, 0.825 ± 0.066, 0.820 ± 0.065, 0.813 ± 0.061, and 0.807 ± 0.060. The BSS scores are 0.491 ± 0.114, 0.494 ± 0.110, 0.495 ± 0.112, 0.492 ± 0.101, and 0.497 ± 0.106.}
	
	{Finally, only the FN reward is changed, while the TP, TN, and FP rewards remain unchanged. Table \ref{tab13} shows the effects of applying a perturbation (step size=1) to the FN reward on the TSS and BSS scores of the CDR-Transformer-10 model. As shown in Table \ref{tab13}, we successively decrease the FN reward from its benchmark in step size of 1. The TSS scores are 0.801±0.050, 0.800±0.072, 0.802±0.056, 0.806±0.070, and 0.811±0.063. The BSS scores are 0.490±0.123, 0.493±0.124, 0.489±0.123, 0.487±0.130, and 0.485±0.135. When the FN reward increases by step size of 1 at a time compared to the benchmark reward value, the TSS scores are 0.813±0.070, 0.813±0.061, 0.807±0.056, 0.819±0.051, and 0.801±0.072. The BSS scores are 0.486±0.113, 0.485±0.107, 0.493±0.110, 0.483±0.122, and 0.498±0.096.}
	
\begin{figure*}
		\centering
		\includegraphics[width=0.92\textwidth]{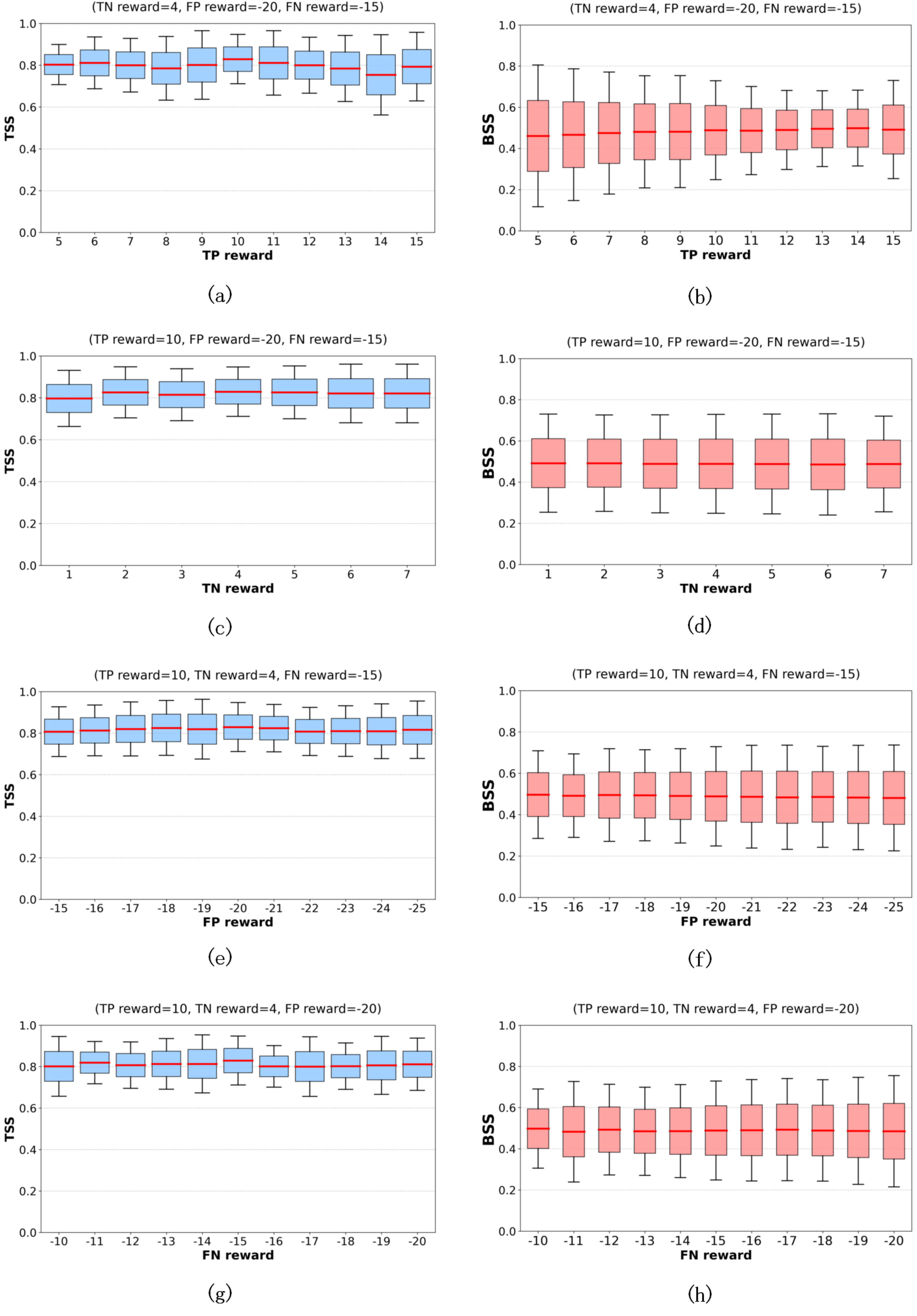}
		\caption{{Boxplots of the effect of reward variation on TSS and BSS of the CDR-Transformer-10 model. In the boxplots, the red horizontal line in each box represents the median value of TSS or BSS.}}\label{figure12}
	\end{figure*}

	{Based on Tables \ref{tab10}, \ref{tab11}, \ref{tab12}, and \ref{tab13}, we use TP reward, TN reward, FP reward, and FN reward as the horizontal axes, with the TSS and BSS values of the model as the vertical axes, respectively, to create boxplots. Figure \ref{figure12} shows the boxplots of the effect of reward variation on TSS and BSS of the CDR-Transformer-10 model. In Figure \ref{figure12}(a), the TP reward changes sequentially from 5 to 15. In Figure \ref{figure12}(c), the TN reward changes from 1 to 7. In Figure \ref{figure12}(e), the FP reward changes from -15 to -25, and in Figure \ref{figure12}(g), the FN reward changes from -10 to -20. The TSS scores of the model remain stable around 0.8, indicating that the TSS scores of the CDR-Transformer-10 model are less affected by changes in TP, TN, FP, and FN rewards. As shown in Figure \ref{figure12}(b), the TP reward changes sequentially from 5 to 15. In Figure \ref{figure12}(d), the TN reward changes sequentially from 1 to 7. In Figure \ref{figure12}(f), the FP reward changes sequentially from -15 to -25. In Figure \ref{figure12}(h), the FN reward changes sequentially from -10 to -20. The BSS scores remain stable around 0.5, indicating that the BSS scores of CDR-Transformer-10 are less affected by the changes in TP, TN, FP, and FN rewards. Therefore, the categorical and probabilistic prediction performance of the CDR-Transformer-10 model is not overly sensitive to this change compared to the benchmark reward value. In addition, we conduct a sensitivity analysis on immediate reward engineering for the CDR-CNN model and the CDR-CNN-BiLSTM model. The TP, TN, FP, and FN rewards also have a relatively small impact on the TSS and BSS scores of the CDR-CNN model and the CDR-CNN-BiLSTM model. Therefore, the prediction performance of CDR-CNN and CDR-CNN-BiLSTM is not overly sensitive to the specific choices of immediate reward values. Overall, the prediction performance of the CDR models (CDR-Transformer-10, CDR-CNN, and CDR-CNN-BiLSTM) is not overly sensitive to changes in TP, TN, FP, and FN rewards compared to the benchmark reward value.}
	
	{The experimental results in Sections \ref{subsec:per comprison} and \ref{subsec:sen} demonstrate that the proposed CDR models with knowledge-informed features not only achieves slightly better performance than the deep learning models but also effectively treats the class imbalance problem through differentiated immediate rewards. Furthermore, this framework is not overly sensitive to the specific choices of immediate reward values. Even if the immediate reward value is perturbed, CDR models can still maintain excellent prediction performance, avoiding the limitations of traditional supervised learning methods that rely on tuning hyperparameters, such as class weights, to alleviate class imbalance problem. }

	\subsection{Model Interpretability Analysis} \label{SHAP}
	
	According to the performance comparison results in Section \ref{subsec:per comprison}, among the deep learning models utilizing knowledge-informed features, Transformer-10 performs best in both categorical and probabilistic forecasting tasks. Similarly, among the {CDR} models utilizing knowledge-informed features, {CDR-Transformer-10} demonstrates the best performance in categorical and probabilistic forecasting. Therefore, Section \ref{SHAP} will focus on a comparative study of model interpretability for deep learning and {CDR} models based on knowledge-informed features.
	
	In Section \ref{SHAP}, we employ the SHAP method to analyze the contributions of 10 knowledge-informed features to the output probabilities of Transformer-10 and {CDR-Transformer-10}. SHAP values can reveal the specific impact of each feature on the model’s predictions, providing an effective way to interpret the results of machine learning models. SHAP values can be either positive or negative. A positive value indicates that the feature increases the model's predicted probability, while a negative value suggests that the feature decreases the predicted probability. The absolute magnitude of the SHAP value reflects the strength of influence that the knowledge-informed feature has on the flare prediction probability. This method allows us to decompose the model’s output $f(x)$ into the sum of contributions from the 10 features over $T$ time steps, expressed as:
	
	\begin{equation}
		f(x) = \phi_0 + \sum_{i=1}^{10} \sum_{t=1}^{T} \phi_{i,t},
		\label{eq:12}
	\end{equation}
	here, $x$ represents the feature vector for an AR; $ \phi_{i,t} $ denotes the contribution of feature $i$ at time step $t$; $T$=40 represents the total number of time steps. And $ \phi_{0} $ is the average predicted value of model $f$ on the training set, also referred to as the base value. For the SHAP value $ \phi_{i,t} $ of feature $i$ at time step $t$, considering a subset $S$ composed of $ |S| $ features, where $ S \subseteq F, (|F|=10) $, the SHAP value of feature $i$ can be expressed as:
	
	\begin{equation}
		\phi_{i,t} = \sum_{S \subseteq F \setminus \{i\}} \frac{|S|!(|F| - |S| - 1)!}{|F|!} \cdot (f(S \cup \{i\}, t) - f(S, t)), \quad (i = 1, \ldots, |F|),
		\label{eq:13}
	\end{equation}
	here, $ f(S \cup \{i\}, t) $ and $f(S,t)$ represent the model's output probabilities at time step $t$ when feature $i$ is included or excluded in the feature combination, respectively. $ \frac{|S|!(|F| - |S| - 1)!}{|F|!} $ represents that when the total number of features is $|F|$, there are a total of $|F|!$ possibilities under ordered permutations; when feature $i$ is fixed, there are $ |S|!(|F| - |S| - 1)! $ possible combinations.
	
	To calculate the global SHAP mean value for feature $i$ across all ARs on the testing set, the SHAP value of feature $i$ are first summed over all time steps for all ARs. The absolute value of this sum is then taken and averaged across all ARs, yielding the global SHAP mean for feature $i$, which can be expressed as:
	
	\begin{equation}
		\phi_i^{global} = \frac{1}{N} \sum_{n=1}^{N} |\phi_i^{(n)}|,
		\label{eq:14}
	\end{equation}
	
	\begin{equation}
		\phi_i^{(n)} = \sum_{t=1}^{T} \phi_{i,t}^{(n)},
		\label{eq:15}
	\end{equation}
	where $N$ is the number of ARs on the testing set; $ \phi_{i,t}^{(n)} $ represents the SHAP value of feature $i$ at time step $t$ for the $n$th AR; and $ \phi_{i}^{(n)} $ denotes the sum of the SHAP values of feature $i$ across all time steps for the $n$th AR.
	
\begin{figure*}
		\centering
		\includegraphics[width=0.7\textwidth]{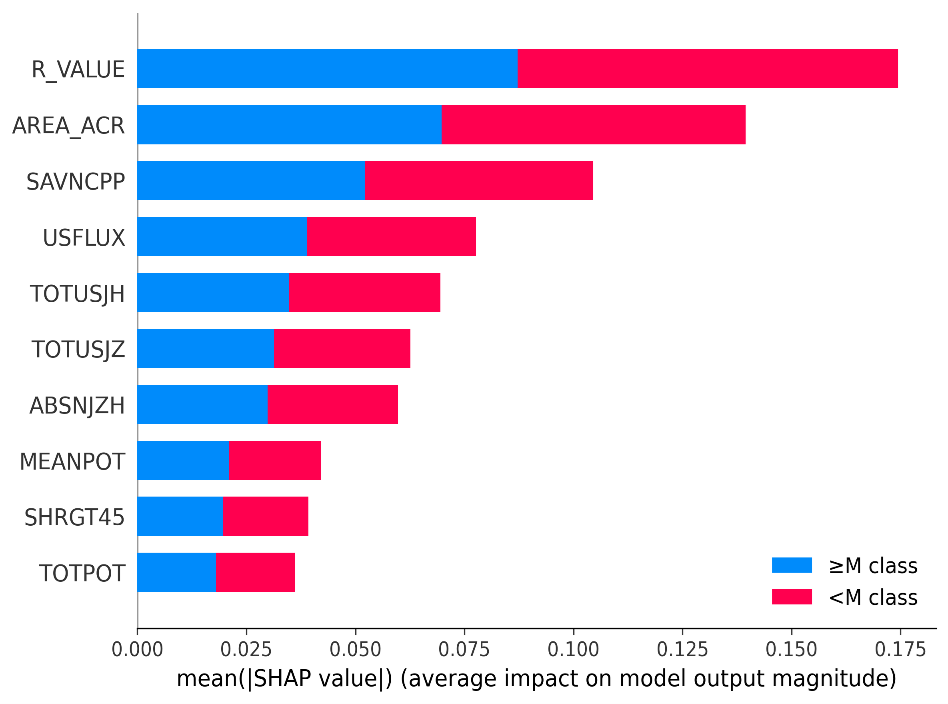}
		\caption{A bar chart depicting the global significance of the 10 features for Transformer-10 on one testing set from 10 CV datasets.}\label{figure3}
	\end{figure*}

\begin{figure*}
		\centering
		\includegraphics[width=0.7\textwidth]{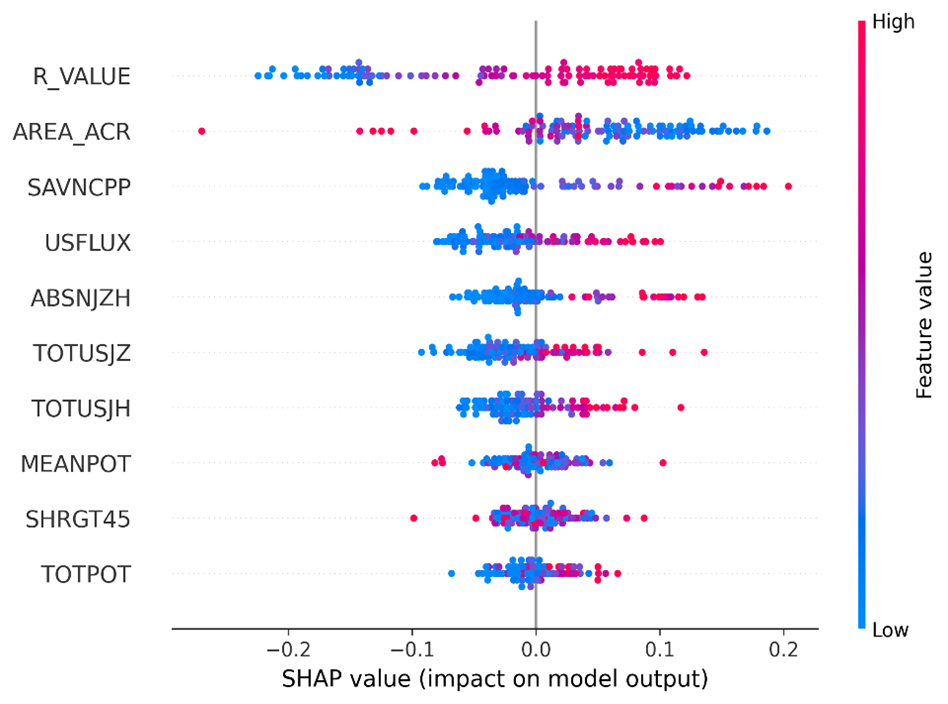}
		\caption{Beeswarm plot showing the effect of the 10 features on Transformer-10 for each AR.}\label{figure4}
	\end{figure*}

	In deep learning based on knowledge-informed features, Figure \ref{figure3} depicts a bar chart of the global significance of the 10 features for Transformer-10 on one testing set from 10 CV datasets. The x-axis represents the global mean SHAP values of each feature calculated via Formula \eqref{eq:14}, while the y-axis lists the 10 features ranked in descending order of SHAP magnitude. As shown in Figure \ref{figure3}, R\_VALUE exhibits the highest mean SHAP value among all features, indicating its dominant influence on the Transformer-10 model for flare predictions. Conversely, SHRGT45 and TOTPOT demonstrate relatively low mean SHAP values, suggesting their minimal impact on the model's forecasting performance. Figure \ref{figure4} presents a beeswarm plot illustrating the effect of the 10 features on Transformer-10 for each AR. The x-axis denotes the SHAP values of any given feature across all time steps for an AR, computed using Formula \eqref{eq:15}, with scatter point colors reflecting the relative magnitude of the feature values. As shown in Figure \ref{figure4}, the plot reveals that as R\_VALUE feature values increase, their corresponding SHAP values also rise, demonstrating a positive promotional effect on the model's output probability. Conversely, decreasing R\_VALUE values lead to reduced SHAP values, highlighting a negative inhibitory effect. Notably, the SHAP values of the R\_VALUE feature for all ARs show a significant clustering distribution away from the vertical line at SHAP value=0, indicating that this feature has a substantial impact on the model's predicted probability. Conversely, the SHAP values of features like SHRGT45 and TOTPOT exhibit a clustering distribution around the vertical line at SHAP value=0, which implies a relatively smaller effect on the model's output probabilities.
	
\begin{figure*}
		\centering
		\includegraphics[width=0.7\textwidth]{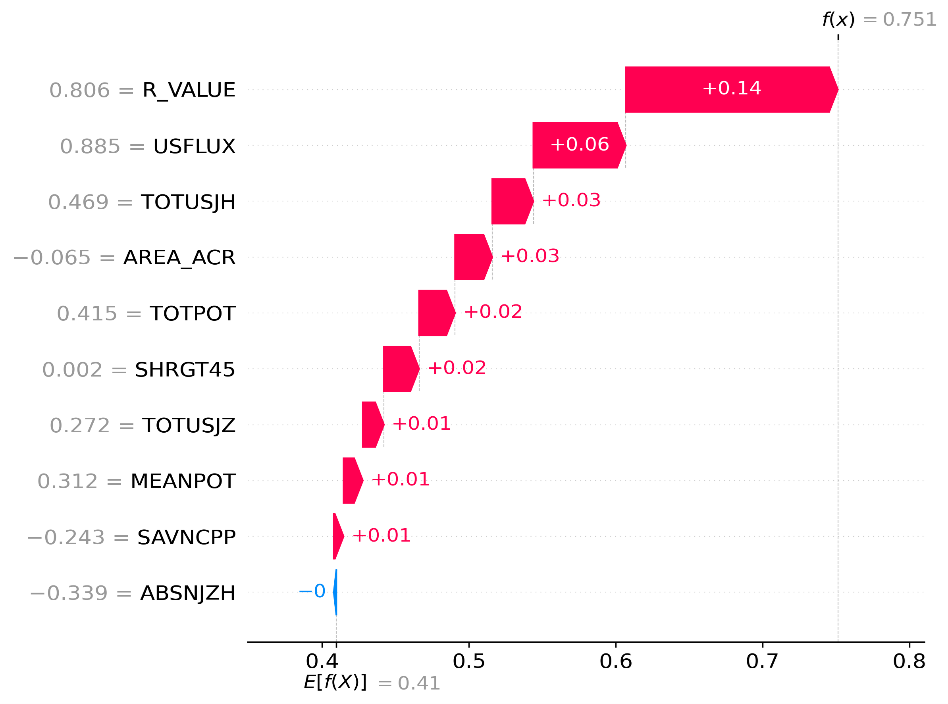}
		\caption{The waterfall plot for the correct prediction of a positive class on Transformer-10 for AR12257. The AR12257 produced an M-class flare at 04:13 UTC on January 13, 2015.}\label{figure5}
	\end{figure*}

\begin{figure*}
		\centering
		\includegraphics[width=0.7\textwidth]{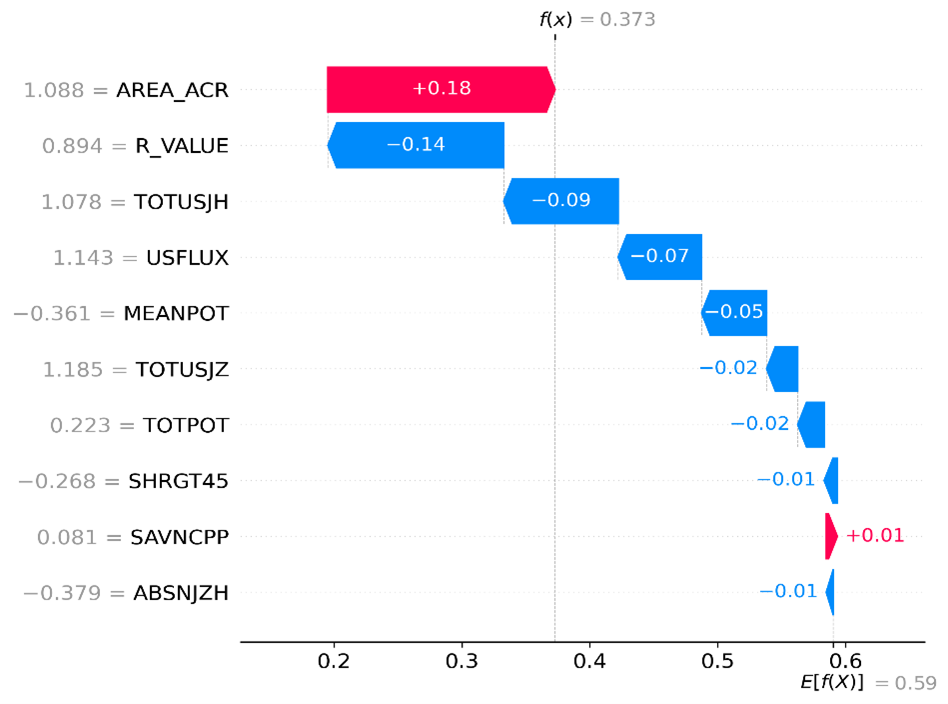}
		\caption{The waterfall plot for the correct prediction of a negative class on Transformer-10 for AR12367. The AR12367 produced a C-class flare at 19:07 UTC on June 20, 2015.}\label{figure6}
	\end{figure*}

	To clearly illustrate how the 10 features influence the model's probability output, we randomly select an AR correctly predicted as a positive class (AR 12257) and one correctly predicted as a negative class (AR 12367). We then plot the SHAP waterfall plots for these two ARs, as shown in Figures \ref{figure5} and \ref{figure6}. In the figures, red indicates that a feature has a positive impact on the model's probability output, while blue indicates a negative impact. The values within the arrow boxes represent the sum of the SHAP values for each feature across 40 time steps, calculated using Formula \eqref{eq:15}. The y-axis lists the 10 features, sorted from highest to lowest SHAP value for that AR. In the figures, $E$[$f(x)$] is jointly determined by $ \phi_0 $ and the sum of the SHAP values for each feature over 40 time steps from a single AR. $f(x)$ denotes the final probability output of the model for that AR. In Figure \ref{figure5}, the R\_VALUE feature increases the model's probability output by 0.14 relative to the baseline $E$[$f(x)$], thereby contributing to a positive class prediction. Among the 10 features, R\_VALUE has the largest positive impact on the probability output for AR 12257. In Figure \ref{figure6}, the R\_VALUE feature decreases the model's probability output by 0.14 relative to the baseline $E$[$f(x)$], thereby facilitating the prediction of the negative class. Among the 10 features, R\_VALUE has the largest negative impact on the probability output for AR 12367. In summary, through the SHAP interpretability analysis, we have demonstrated the influence of each feature on the model's final output from both global and local perspectives. Among the 10 knowledge-based features in the deep learning model, R\_VALUE plays a crucial role in determining whether the model can correctly predict solar flare occurrence.
	
\begin{figure*}
		\centering
		\includegraphics[width=0.7\textwidth]{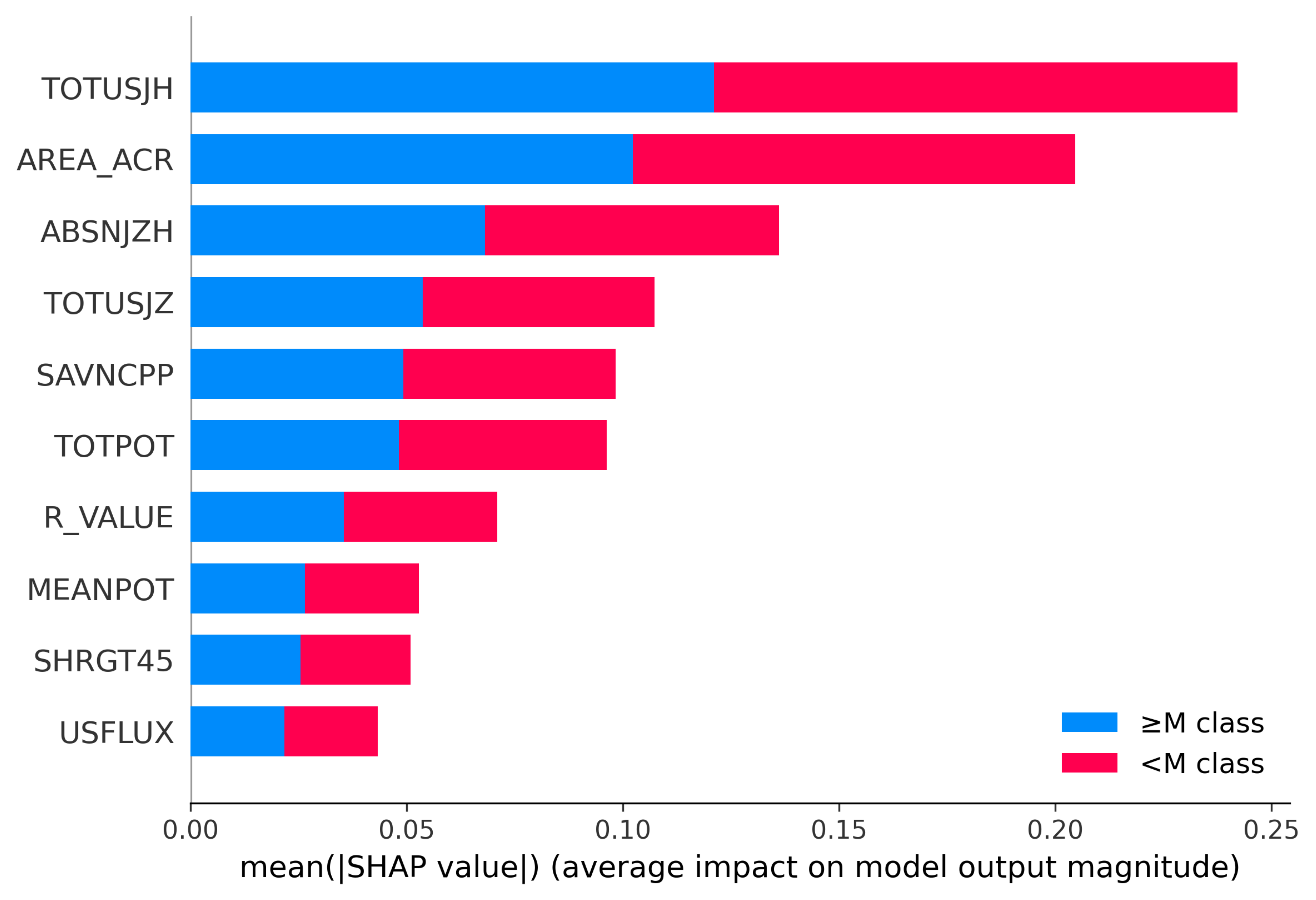}
		\caption{{A bar chart depicting the global significance of the 10 features for CDR-Transformer-10 on one testing dataset from 10 CV datasets.}}\label{figure7}
	\end{figure*}

\begin{figure*}
		\centering
		\includegraphics[width=0.7\textwidth]{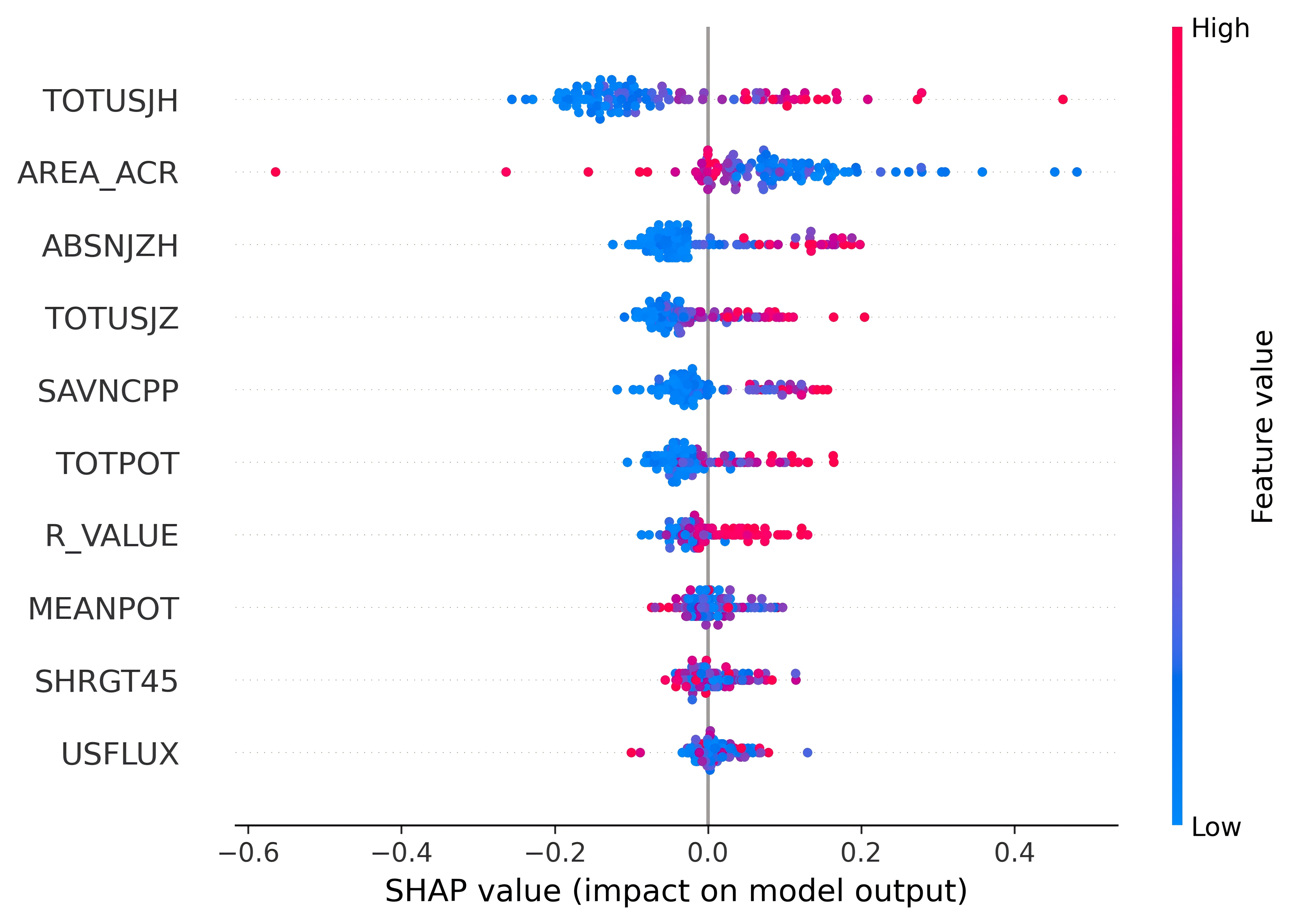}
		\caption{{Beeswarm plot showing the effect of the 10 features on CDR-Transformer-10 for each AR.}}\label{figure8}
	\end{figure*}

\begin{figure*}
		\centering
		\includegraphics[width=0.7\textwidth]{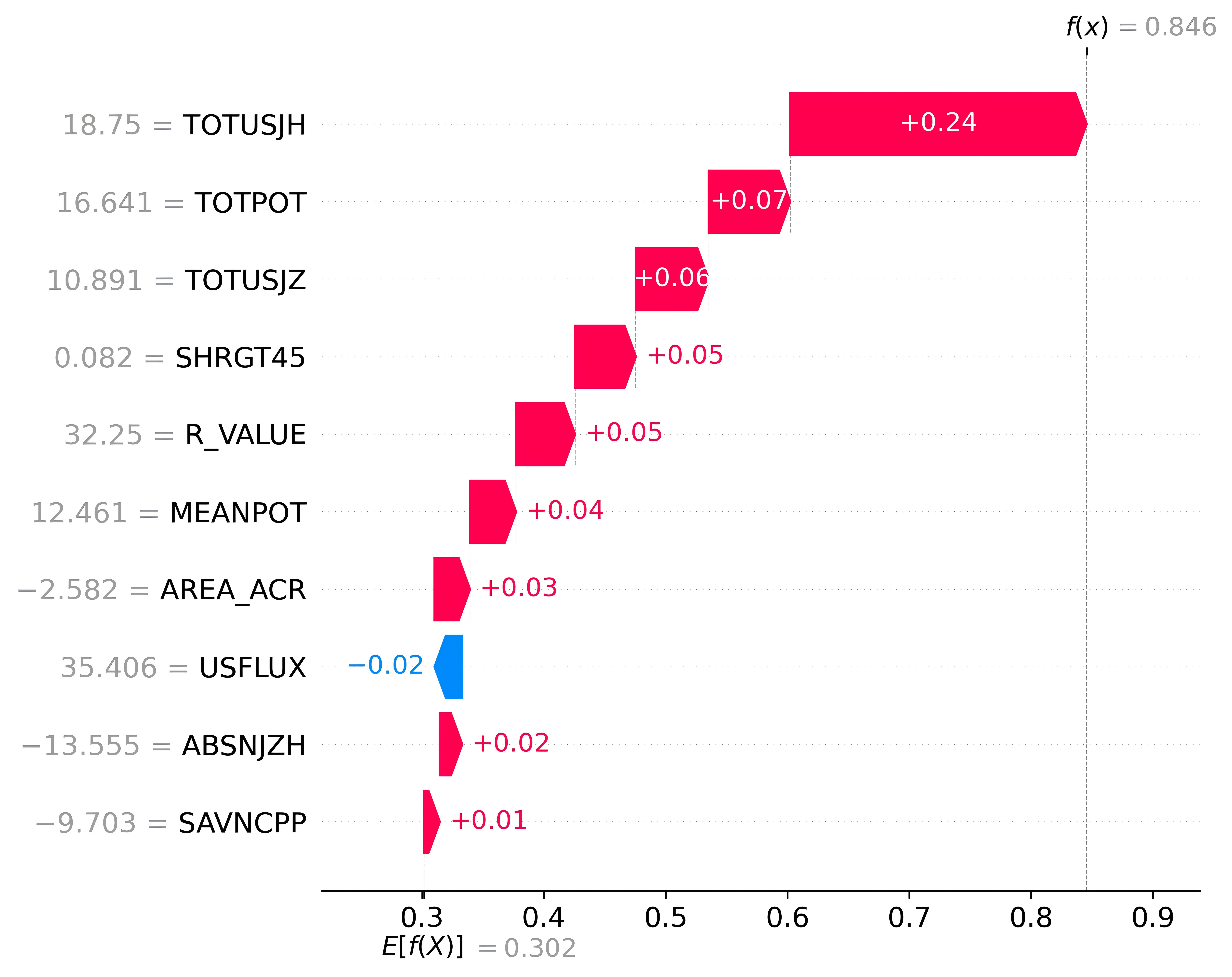}
		\caption{{The waterfall plot for the correct prediction of a positive class on CDR-Transformer-10 for AR12257. The AR12257 produced an M-class flare at 04:13 UTC on January 13, 2015.}}\label{figure9}
	\end{figure*}

\begin{figure*}
		\centering
		\includegraphics[width=0.7\textwidth]{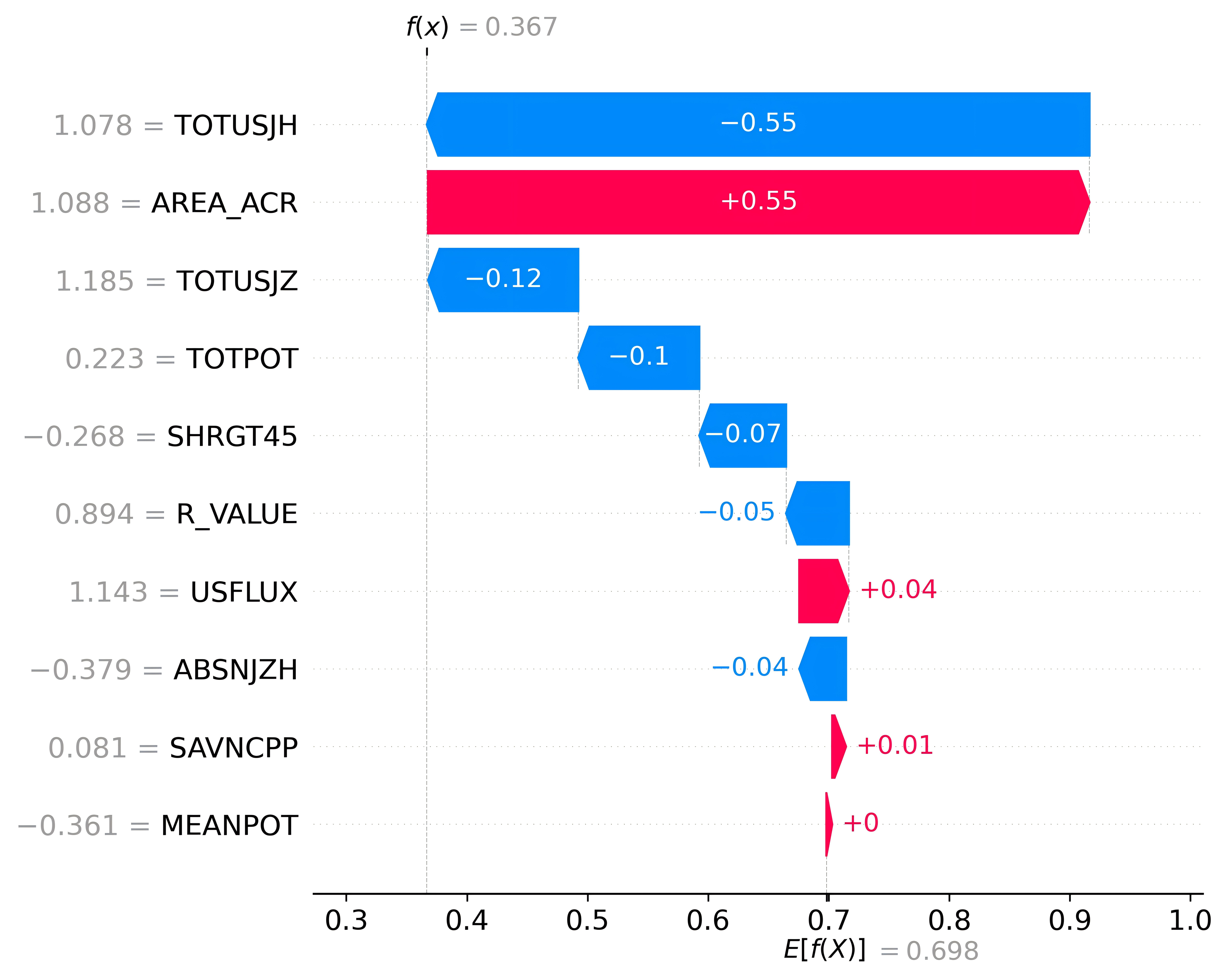}
		\caption{{The waterfall plot for the correct prediction of a negative class on CDR-Transformer-10 for AR12367. The AR12367 produced a C-class flare at 19:07 UTC on June 20, 2015.}}\label{figure10}
	\end{figure*}

	To ensure a fair comparison of SHAP interpretability, both the {CDR} model ({CDR-Transformer-10}) and the deep learning model (Transformer-10) use the same testing set. Figures \ref{figure7} to \ref{figure10} are generated in a manner similar to Figures \ref{figure3} to \ref{figure6}. Figure \ref{figure7} displays a bar chart depicting the global significance of the 10 features for {CDR-Transformer-10} on one testing dataset from 10 CV datasets. As shown in Figure \ref{figure7}, TOTUSJH has the highest mean SHAP value among all features, indicating that it exerts the greatest influence on the model's output probability. In contrast, {SHRGT45} and USFLUX exhibit smaller mean SHAP values, suggesting their relatively minor impact on flare forecasting. Figure \ref{figure8} presents a beeswarm plot illustrating the effect of the 10 features on {CDR-Transformer-10} for each AR. As depicted in Figure \ref{figure8}, when the TOTUSJH feature value increases, its corresponding SHAP value rises, demonstrating a positive contribution to the model's output probability. Conversely, when the TOTUSJH feature value decreases, the SHAP value declines, reflecting a negative inhibitory effect. The SHAP values of the TOTUSJH feature for all ARs show a significant clustering distribution away from the vertical line at SHAP value= 0, indicating that this feature has a substantial impact on the model's predicted probability. In contrast, the SHAP values of features such as {SHRGT45} and USFLUX exhibit a clustering distribution near the vertical line at SHAP value= 0, which implies a comparatively smaller effect on the model's output probabilities.
	
	To better illustrate how the 10 features influence the model's output probability, and to ensure a fair comparison, we select the two ARs with the same numbering as those analyzed in the interpretability study of the deep learning model (Transformer-10). One AR is correctly predicted as a positive class (i.e., AR 12257) and the other is correctly predicted as a negative class (i.e., AR 12367). We then plot SHAP waterfall plots for these two ARs, as shown in Figures \ref{figure9} and \ref{figure10}. In Figure \ref{figure9}, the TOTUSJH feature increases the model's probability output value by {0.24} compared to the baseline value $E$[$f(x)$], thus supporting the prediction of the positive class. Among the 10 features, TOTUSJH has the largest positive impact on the model's probability output for AR 12257. In Figure \ref{figure10}, the TOTUSJH feature decreases the model's probability output value by {0.55} compared to the baseline value $E$[$f(x)$], thus contributing to a negative class prediction. Among the 10 features, TOTUSJH has the largest negative impact on the model's probability output for AR 12367.
	
	Through the SHAP interpretability analysis, we demonstrate the influence of various features on the model's final output from both global and local perspectives. Among the 10 knowledge-based features in {CDR models}, TOTUSJH has the most significant impact on whether the model correctly predicts flare occurrence. However, when performing SHAP interpretability analysis on deep learning {models} and {CDR} models using the same testing set, we observe differences in the results of feature importance. This significant divergence in feature importance analysis between deep learning and {CDR models} likely stems from fundamental differences in their training objectives and optimization processes. {Transformer} directly optimizes the difference between predicted results and true labels by minimizing the loss function, as shown in Equation \eqref{eq:2}. {However, CDR-Transformer optimizes the online network Q using Formula \eqref{eq:4}, thereby achieving the maximum immediate reward. It is worth noting that the immediate reward value is assigned independently based on the prediction result for each AR during the training process, and the model also considers the temporal relationships of the 40 time series samples contained in each AR.} {Furthermore, from the grounded physical interpretation, the parameter TOTUSJH means the total unsigned current helicity, which correlates with the AR size and is thought to be "extensive parameter" \citep{Welsch_2009}. It reflects the long-term process of energy build-up (high correlation between TOTUSJH and $E_{free}$ in Figure 4d of  \citealt{Li_2024}) and is more stable during the AR evolution (energy evolution in Figure 3 of \citealt{2023AdSpR..71.2017K}). However, R\_VALUE is the total unsigned magnetic flux in a 15 Mm-wide region along the strong-field, high-gradient PIL \citep{schrijver2007characteristic}. It varies a lot during the AR evolution (R\_VALUE variation in Figure 3 of \citealt{2023AdSpR..71.2017K}) and is more immediate, which is probably influenced by the shear motion or flux emergence nearby the PILs. Thus, the feature importance difference between the Transformer and CDR-Transformer may be caused by the different characteristics between TOTUSJH and R\_VALUE. The reward scheme in CDR-Transformer might favour the long-term and stable TOTUSJH. Therefore, Transformer and CDR-Transformer focus on feature capabilities from different perspectives, leading to different feature contribution allocations during SHAP calculation.}

	\subsection{Performance Comparison with NASA/CCMC} \label{subsec:NASA}
	
	The CCMC, primarily operated by NASA Goddard Space Flight Center, is dedicated to the development of state-of-the-art space science models aimed at supporting the space weather research community \citep{hesse2001collisionless}. NASA/CCMC has integrated a range of advanced solar flare forecasting models. In this study, we collect real-time solar flare prediction data from various NASA/CCMC models for the period spanning April 1, 2023, to July 21, 2024, obtained from its official website (\url{https://ccmc.gsfc.nasa.gov/scoreboards/flare/}). To compare the performance of the deep learning and {CDR} models (i.e., Transformer-10 and {CDR-Transformer-10}) selected from Section \ref{subsec:per comprison} with that of NASA/CCMC and \citet{111}, we construct a new SHARP testing dataset covering the period from April 1, 2023, to July 21, 2024. The data collection methodology for this testing dataset aligns with the approach described in Section \ref{sec:Data}. This testing dataset is divided into two subsets, namely the original testing dataset and the filtered testing dataset. The prediction time and AR number within these two testing datasets are consistent with those  from NASA/CCMC and \citet{111}. The original testing dataset includes knowledge-informed features for both single and multiple ARs, encompassing 86 ARs distributed as follows: 29 with No-flare class ARs, 40 with C-class ARs, 14 with M-class ARs, and 3 with X-class ARs. The filtered testing dataset, which includes only knowledge-informed features from single AR, consists of 39 ARs, distributed as follows: 16 with No-flare class ARs, 18 with C-class ARs, 3 with M-class ARs, and 2 with X-class ARs. The Transformer-10 and {CDR-Transformer-10} models, previously trained as described in Section \ref{subsec:per comprison}, are tested on both the original and filtered testing datasets. The testing results are compared with the performance of other models (e.g., NASA/CCMC) to evaluate their predictive capabilities.
	
\begin{figure*}
		\centering
		\includegraphics[width=0.8\textwidth]{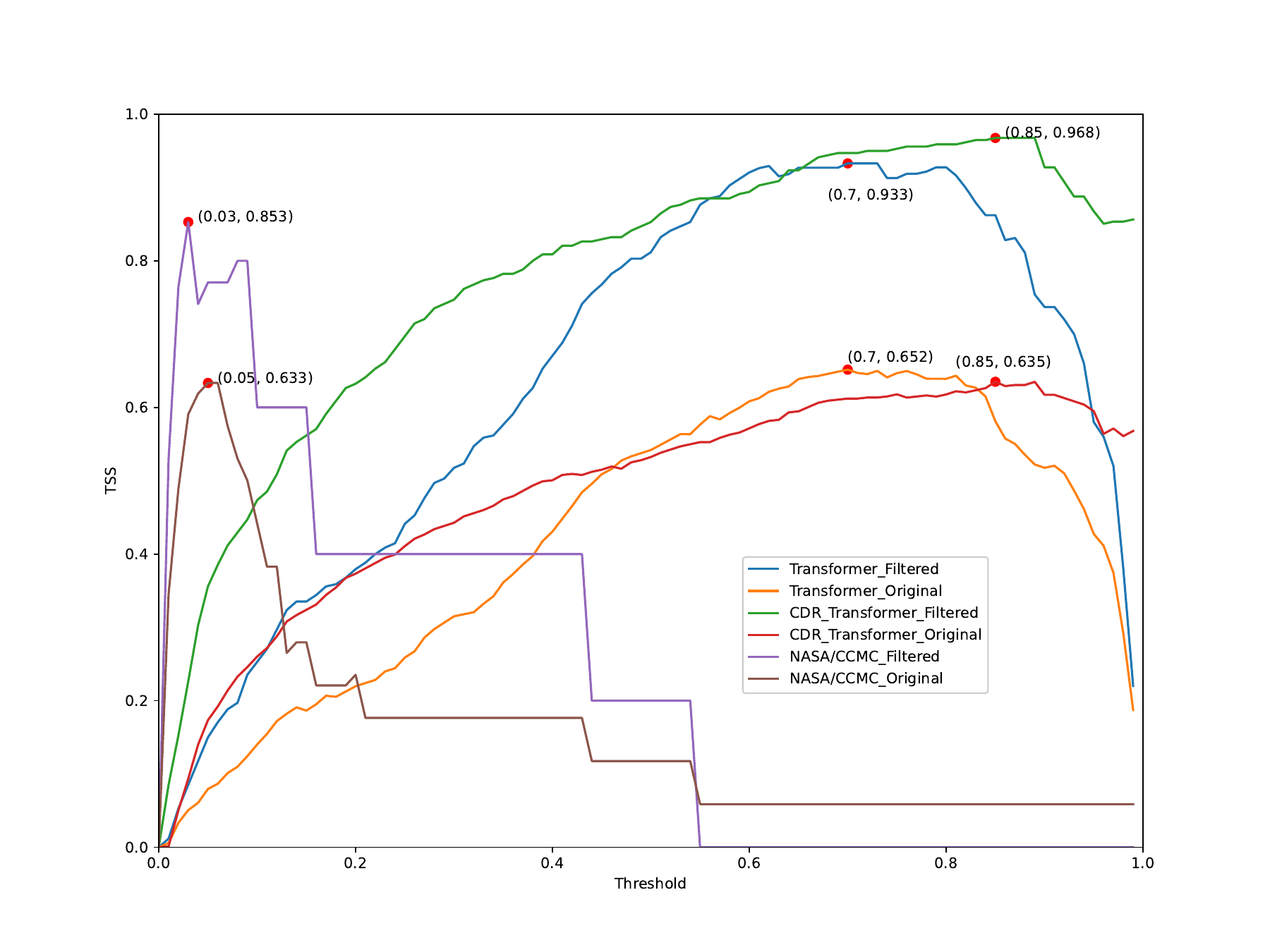}
		\caption{{The variation curves of TSS for Transformer-10, CDR-Transformer-10, and NASA/CCMC across different probability thresholds.}}\label{figure11}
	\end{figure*}

	We adopt a global threshold scanning method by setting a probability threshold to convert the output probabilities of the model into predicted categories. Specifically, the probability threshold is iterated from 0\% to 100\% in 1\% increments. For each threshold, the corresponding TSS value is calculated, enabling a comprehensive evaluation for the performance of the model. Figure \ref{figure11} illustrates the variation curves of TSS for Transformer-10, {CDR-Transformer-10}, and NASA/CCMC across different probability thresholds. The TSS curves for Transformer-10 and {CDR-Transformer-10} are derived from evaluations using the 10 models trained in Section \ref{subsec:per comprison}  on both the original and filtered testing datasets, with the average of the 10 TSS scores calculated for each threshold. The NASA/CCMC covers multiple flare forecasting models, with multiple prediction probabilities generated daily for each AR. However, not all models from NASA/CCMC provide forecasting probability values for ARs, resulting in a relatively small amount of forecasting probability data collected for each model. Consequently, in Section \ref{subsec:NASA}, we utilize the average prediction probabilities from multiple NASA/CCMC models for each AR.
	
	As depicted in Figure \ref{figure11}, the TSS scores for the original and filtered testing datasets initially increase and then decrease as the probability threshold increases. On the original testing dataset, {CDR-Transformer-10}, Transformer-10, and NASA/CCMC achieve their maximum TSS scores at probability thresholds of {85\%}, 70\%, and 5\%, respectively. On the filtered testing dataset, the maximum TSS scores for {CDR-Transformer-10}, Transformer-10, and NASA/CCMC occur at probability thresholds of {85\%}, 70\%, and 3\%, respectively. Table \ref{tab14} shows the performance results for {CDR-Transformer-10}, Transformer-10, iTransformer \citep{111}, and NASA/CCMC at the probability thresholds corresponding to their respective optimal TSS. The TSS scores for {CDR-Transformer-10}, Transformer-10, iTransformer, and NASA/CCMC on the original testing dataset are {0.635 ± 0.045}, 0.652 ± 0.060, 0.668±0.060, and 0.633, respectively, while on the filtered testing dataset, the TSS scores increase to {0.968 ± 0.016}, 0.933 ± 0.057, 0.906±0.043, and 0.853, respectively. Based on the experimental results, on the original testing dataset, the TSS scores of our {CDR-Transformer-10} and Transformer-10 models are comparable to those of the iTransformer model. On the filtered testing dataset, the TSS scores of both our {CDR-Transformer-10} and Transformer-10 models surpass those of the iTransformer model. Additionally, the TSS scores of {CDR-Transformer-10} and Transformer-10 exceed those of NASA/CCMC. On the filtered testing dataset, the TSS scores of {CDR-Transformer-10 } and Transformer-10 are significantly higher than those of NASA/CCMC, with {CDR-Transformer-10} performing the best. Moreover, the models evaluated on the filtered testing dataset demonstrate superior forecasting performance compared to those on the original testing dataset. This difference is primarily attributed to the original testing dataset potentially including multiple ARs of different classes, which may affect the computation of knowledge-informed features and thereby reduce prediction performance of the model.

	\section{Conclusions and discussions} \label{sec:conclu}
	
	In this study, {for the first time, we develop a CDR framework to predict $\geq$M class flares within 24 hr. We} construct multiple datasets encompassing knowledge-informed features and magnetogram images. We {also} develop three deep learning models (i.e., CNN, CNN-BiLSTM, and Transformer) and three {CDR} models (i.e., {CDR-CNN, CDR-CNN-BiLSTM, and CDR-Transformer}) to perform both categorical and probabilistic predictions for $\geq$M-class solar flares within 24 hr. The architectures of the deep learning and {CDR} models are identical. We primarily utilize TSS and BSS metrics to evaluate the categorical and probabilistic prediction performance of these models, respectively. {First}, we perform an importance analysis of 31 LOS magnetic field parameters using the Transformer model. {In addtion}, we compare the categorical and probabilistic prediction performance of LOS magnetic field parameters, vector magnetic field parameters, and the combination of both using the Transformer model. {Second}, we compare the performance of the flare prediction based on {CDR models} ({CDR-CNN, CDR-CNN-BiLSTM, and CDR-Transformer}) and {deep learning models} ({CNN, CNN-BiLSTM, and Transformer}). {Third, we conduct a sensitivity analysis on reward engineering for CDR models (CDR-Transformer-10, CDR-CNN, CDR-CNN-BiLSTM).} {Fourth, we use the SHAP method to conduct interpretability research on both the deep learning model (Transformer) and the CDR model (CDR-Transformer).} Finally, under identical prediction time and AR number, we conduct a comparative study of forecasting performance of our models with that of other models (e.g., NASA/CCMC).
	
	The main results of this study are summarized as follows: (1) Among various combinations of LOS knowledge-informed features, the combination of R\_VALUE and AREA\_ACR exhibits the best performance in both categorical and probabilistic forecasting for the knowledge-informed model (i.e., Transformer). (2) The Transformer demonstrates superior prediction performance when utilizing the combination of LOS and vector magnetic field parameters compared to using either LOS or vector magnetic field parameters alone. (3) In both categorical and probabilistic predictions, the models based on knowledge-informed features (i.e., Transformer-10, Transformer-2, {CDR-Transformer-10}, and {CDR-Transformer-2}) outperform those based on magnetogram images (i.e., CNN, CNN-BiLSTM, {CDR-CNN}, and {CDR-CNN-BiLSTM}). (4) The deep learning models (CNN and CNN-BiLSTM) outperforms the {CDR} counterparts ({CDR-CNN, CDR-CNN-BiLSTM}) when both are based on magnetogram images, while the {CDR} models ({CDR-Transformer-10 and CDR-Transformer-2}) {are slightly superior to} the deep learning counterparts (Transformer-10 and Transformer-2) when both utilize knowledge-informed features. Among all the models, the {CDR-Transformer-10} achieves the highest prediction performance. {(5) We conduct immediate reward value perturbations for TN, TP, FN, and FP rewards, respectively. The changes in the TSS and BSS scores of the model are relatively small, indicating that the prediction performance of the CDR model (CDR-Transformer-10, CDR-CNN, and CDR-CNN-BiLSTM) is not overly sensitive to the specific choices of immediate reward values.} {(6)} Through global and local SHAP analysis, we identify the contribution of knowledge-informed features to the output probability of both deep learning and {CDR} models. R\_VALUE is the most crucial feature for flare prediction in the deep learning model, while TOTUSJH holds the most importance for flare prediction in the {CDR} model. {(7)} The {CDR} model ({CDR-Transformer-10}) and deep learning model (Transformer-10), both utilizing 10 knowledge-informed features, achieve TSS scores of {0.968 ± 0.016} and 0.933 ± 0.057, respectively, on the filtered testing dataset. Our models demonstrate superior prediction performance compared to NASA/CCMC.

	In the performance comparison between our models and NASA/CCMC, NASA/CCMC employs multiple methods, including ASAP \citep{abed2021automated}, MAG4 \citep{falconer2011tool}, and NOAA \citep{crown2012validation}. ASAP applies CNNs to predict solar flares based on sunspots McIntosh classification, using SDO/HMI Intensitygram images and magnetogram images. Meanwhile, MAG4 predicts the event occurrence rate for each AR by measuring a free-energy proxy from LOS or vector magnetogram images. The NOAA, rooted in climatological methods, classifies ARs and assigns probabilities based on the historical flare rates of different sunspot region classes. Although NASA/CCMC employs relatively robust flare prediction algorithms, our study leverages more advanced deep learning techniques (i.e., Transformer) combined with reinforcement learning, to further enhance flare prediction accuracy. The experimental results demonstrate that {CDR-Transformer-10} is the most effective model for predicting whether $\geq$M-class flares will occur within 24 hr.
	
	{Given that the CDR model not only achieves excellent prediction performance by incorporating knowledge-informed features but also effectively treats the class imbalance problem, it is better suited to meet the practical demands of space weather forecasting. Therefore, our future work will focus on developing an operational solar flare prediction system based on the CDR model, with the goal of providing accurate and real-time flare forecasts. This system is expected to make significant contributions to global space weather forecasting. We will also apply the CDR framework to the forecasting of coronal mass ejections (CMEs), which occur far less frequently than flares \citep{raju2023interpretable}. In addition, with the increasing number of solar observation satellites, such as the Advanced Space-based Solar Observatory \citep[{ASO};][]{gan2025aso}, the volume of solar observation data continues to grow rapidly. The soon-to-be-launched Lagrange-V Solar Observation {(LAVSO; \citealt{xihe2}, also known as "Xihe-2")} and the Solar Polar orbit Observatory {(SPO; \citealt{deng2025probing}, also known as "Kuafu-2")} will provide even more solar observation data. Our CDR framework will integrate these multisource data to build more accurate solar activity forecasting models, thereby enhancing the precision of space weather forecasting from its source.}

	\section*{Acknowledgements}
	
	We would like to express our gratitude to the anonymous reviewers for their constructive comments and suggestions, which have greatly enhanced the quality of this paper. The data used in this study is provided by the SDO science teams. This research is supported by the National Natural Science Foundation of China (Grant No. 12473056), the Natural Science Foundation of Jiangsu Province (Grant No. BK20241830), the Qing Lan Project{, and the Specialized Research Fund for State Key Laboratories.}.
	
	\section*{Data Availability}

	The data and code used in this study are available via doi: (\url{https://doi.org/10.5281/zenodo.17928662}).

	
	
	\bibliographystyle{mnras}
	\bibliography{example} 

	
	
	
	\appendix
	
	

\begin{table*}
	\centering
	\caption{31 knowledge-informed features calculated from SHARP LOS magnetogram images.}\label{tab1}%
	\renewcommand{\arraystretch}{1.3} 
	\renewcommand{\arraystretch}{1.5}
	\begin{tabular}{l l m{6cm} l}
		\toprule
		\textbf{Group name of knowledge-informed features } & \textbf{Keyword} & \textbf{Description} & \textbf{Reference} \\ 
		\midrule
		
		\multirow{2}{*}{\shortstack[l]{R\_VALUE \& \\ AREA\_ACR}} 
		& R\_VALUE & Sum of flux near polarity inversion line & \citet{schrijver2007characteristic} \\
		\cline{2-4}
		& AREA\_ACR & Area of strong field pixels in the AR & \citet{bobra2015solar}\\ 
		\midrule
		
		\multirow{7}{*}{Gradient features} 
		& Gradient mean & \multirow{7}{=}{To condense the gradient information into single descriptors (features) for each image, the gradient features are computed as the (1) mean, (2) standard deviation, (3) maximum, (4) minimum, (5) median, (6) skewness, and (7) kurtosis of the gradients in each image.} & \\
		& Gradient std & & \\
		& Gradient median & & \\
		& Gradient min & & \\
		& Gradient max & & \\
		& Gradient skewness & & \\
		& Gradient kurtosis & & \\ 
		\cline{1-3}
		
		\multirow{5}{*}{Wavelet features} 
		& Wavelet energy level 1 & \multirow{5}{=}{Wavelet features characterizing the structure of the magnetic flux at different size scales are extracted by applying a five-level Haar wavelet decomposition.} & \\
		& Wavelet energy level 2 & & \\
		& Wavelet energy level 3 & & \\
		& Wavelet energy level 4 & & \\
		& Wavelet energy level 5 & & \\ 
		\cline{1-3}
		
		\multirow{13}{*}{\shortstack[l]{Neutral line \\ features}} 
		& NL length & The neutral line (NL) length is determined as the sum of the pixels in the strong-gradient binary NL mask obtained by thresholding the gradient-weighted neutral line (GWNL) at 20\% of its maximum value. & \\
		\cline{2-3}
		& NL no. fragments & The number of NL fragments is defined as the number of 8-connected components in the strong-gradient binary NL mask image obtained by thresholding the GWNL at 20\% of its maximum value. & \\
		\cline{2-3}
		& NL gradient-weighted length & The gradient-weighted length of the NL is computed by summing the pixels in the GWNL image. & \\ \cline{2-3}
		& NL curvature mean & \multirow{5}{=}{The curvature angles are computed separately for each NL segment, and the mean, standard deviation, maximum, minimum, and median are computed for all curvature angles for all NL segments.} & \\
		& NL curvature std & & \\
		& NL curvature median & & \\
		& NL curvature min & & \\
		& NL curvature max & & \\ \cline{2-3}
		& NL bending energy mean & \multirow{5}{=}{The bending energy is computed as the normalized sum of the squared difference in curvature between subsequent boundary points, which serves as a measure of the shape of the NL. This measure is computed separately for each NL and the mean, standard deviation, maximum, minimum, and median are computed for the distribution of bending energy.} & \\
		& NL bending energy std & & \\
		& NL bending energy median & & \\
		& NL bending energy min & & \\
		& NL bending energy max & & \\ 
		\cline{1-3}
		
		\multirow{4}{*}{Flux features} 
		& Total unsigned flux & The total unsigned magnetic flux is the absolute sum of the magnetogram image in the AR. & \\
		\cline{2-3}
		& Total signed flux & The total signed magnetic flux is the sum of the magnetogram image in the AR. & \\
		\cline{2-3}
		& Total negative flux & The total negative flux is the sum of negative values of the magnetogram image in the AR. & \\
		\cline{2-3}
		& Total positive flux & The total positive flux is the sum of positive values of the magnetogram image in the AR. 
		& \multirow{-33}{*}{\shortstack{\citet{al2015automated} \\ \citet{boucheron2023solar}}} \\ 
		\bottomrule
	\end{tabular}
\end{table*}

	\begin{table*}
		\centering
		\caption{8 Knowledge-informed features derived from SHARP vector magnetograms.}. \label{tab2}%
		\begin{tabular}{l l l }
			\hline
			Keyword  & Description                                                                                   & Formula                                             \\ \hline
			TOTUSJZ   & Total unsigned vertical current            & $J_{Z_{total}}=\sum \mid J_{Z}\mid dA$ \\
			TOTUSJH   & Total unsigned current helicity                                                               & $H_{C_{total}}\propto \sum \mid B_{Z}\cdot J_{Z} \mid$\\
			TOTPOT    & Total photospheric magnetic free energy density     & $\rho _{tot}\propto \sum \left ( B^{Obs} -B^{Pot}\right )^{2}dA $                                                 \\
			ABSNJZH   &  Absolute value of the net current helicity   & $H_{C_{abs}}\propto \mid \sum B_{Z} \cdot J_{Z}\mid$                                                    \\
			SAVNCPP   &  Sum of the modulus of the net current per polarity & $J_{Z_{sum}}\propto \mid \sum_{}^{B_{Z}^{+}}J_{Z}dA \mid +\mid \sum_{}^{B_{Z}^{-}}J_{Z}dA \mid$                                                    \\
			USFLUX    & Total unsigned flux                                                                           & $\varphi =\sum \mid B_{Z} \mid dA$                                                    \\
			MEANPOT   &  Mean photospheric magnetic free energy        & $\bar{\rho }\propto \frac{1}{N}\sum \left ( B^{Obs}-B^{Pot} \right )^{2}$                                                    \\
			SHRGT45   & Fraction of area with shear \textgreater{}45°     & $Area\ with\ shear > 45^{\circ} / total\_area$                                                  \\ \hline
		\end{tabular}
	\end{table*}

	\begin{table*}
		\centering
		\caption{The distribution of the number of ARs from the 10 CV datasets used in this study, including magnetogram images and knowledge-informed features. The “No-flare /C/M/X” in the table represents the number of ARs corresponding to their respective categories among the training, validation, and testing sets.}
		\label{tab3}
		\begin{tabular}{cccc}
			\hline
			Dataset & Training(No-flare/C/M/X) & Validation(No-flare/C/M/X) & \multicolumn{1}{l}{Testing(No-flare/C/M/X)} \\ \hline
			No.1    & 223/168/46/8              & 70/37/11/1                 & 58/33/16/4                                  \\
			No.2    & 223/168/46/8              & 64/38/12/1                 & 63/37/16/3                                  \\
			No.3    & 223/168/46/8              & 67/37/10/1                 & 59/30/14/4                                  \\
			No.4    & 223/168/46/8              & 66/37/8/2                  & 53/29/17/5                                  \\
			No.5    & 223/168/46/8              & 62/37/10/2                 & 56/30/13/4                                  \\
			No.6    & 223/168/46/8              & 64/33/9/1                  & 50/30/17/3                                  \\
			No.7    & 223/168/46/8              & 65/27/7/2                  & 61/36/16/4                                  \\
			No.8    & 223/168/46/8              & 69/39/9/3                  & 58/33/19/2                                  \\
			No.9    & 223/168/46/8              & 63/45/12/2                 & 60/26/13/3                                  \\
			No.10   & 223/168/46/8              & 69/40/10/2                 & 53/32/17/4                                  \\ \hline
		\end{tabular}
	\end{table*}

	\begin{table*}
		\centering
		\caption{The parameter configuration of deep learning models.} \label{tab4}%
		\begin{tabular}{cccc}
			\hline
			Parameter   name & CNN                       & CNN-BiLSTM                & Transformer               \\ \hline
			Batch size       & 16                        & 16                        & 10                        \\
			Epochs           & 120                       & 120                       & 150                       \\
			Loss function    & cross-entropy & cross-entropy & cross-entropy \\
			Optimizer        & SGD                       & SGD                       & Adam                      \\
			Learning rate    & 0.01                      & 0.01                      & 0.0001                    \\ \hline
		\end{tabular}
	\end{table*}

	\begin{table*}
		\centering
		\caption{The parameter configuration of {CDR} models.}. \label{tab5}%
		\begin{tabular}{cccc}
			\hline
			Parameter   name         & {CDR-CNN}                   & {CDR-CNN-BiLSTM}            & {CDR-Transformer  }         \\ \hline
			TP reward                & {4 }                       & 7                         & 10                        \\
			TN reward                & {10}                         & 4                         & 4                         \\
			FP reward                & {-42 }                      & -24                       & {-20}                       \\
			FN reward                & {-15}                        & -8                        & -15                      \\
			Batch size               & 8                         & 8                         & {49}                        \\
			{Episode}                   & 8                         & 10                        & {8}                       \\
			Loss function            & cross-entropy & cross-entropy & cross-entropy \\
			Optimizer                & Adam                      & Adam                      & Adam                      \\
			Learning rate            & 0.003                     & 0.001                     & 0.00015                   \\
			Exploration decay        & 0.7                       & 0.7                       & 0.99                      \\
			Experience pool size     & 10000                     & 1000                      & 1000                      \\
			 \hline
		\end{tabular}
	\end{table*}

	\begin{table*}
		\centering
		\caption{\label{tab6}%
			The TSS score for categorical prediction of $\geq$M-class flares using the Transformer model, as well as the BSS score for probabilistic prediction, involving a total of 10 groups of LOS knowledge-informed features.}. 
		\begin{tabular}{cccc}
			\hline
			\multirow{2}{*}{Feature}                         & \multirow{2}{*}{Model}        & \multicolumn{2}{c}{Metric}    \\ \cline{3-4} 
			&                               & TSS            & BSS          \\ \hline
			All   features                                   & \multirow{10}{*}{Transformer} & 0.636±0.048    & 0.293±0.067  \\
			R\_VALUE\&AREA                                   &                               & 0.697±0.137    & 0.448±0.079  \\
			Gradient   features                              &                               &0.570± 0.083  & 0.154±0.126 \\
			Neutral   line features                          &                               & 0.347± 0.126  & -0.063±0.129 \\
			Wavelet   features                               &                               & 0.418± 0.097  & -0.109±0.158 \\
			Flux   features                                  &                               & 0.366± 0.105  & -0.447±0.138 \\
			R\_VALUE   \& AREA\_ACR \& Gradient features     &                               & 0.660 ± 0.124 & 0.345±0.094 \\
			R\_VALUE   \& AREA\_ACR \& Neutral line features &                               & 0.664± 0.072  & 0.379±0.089 \\
			R\_VALUE   \& AREA\_ACR \& Wavelet features      &                               & 0.674 ± 0.129 & 0.357±0.109 \\
			R\_VALUE   \& AREA\_ACR \& Flux features         &                               & 0.695 ± 0.080 & 0.349±0.119 \\ \hline
		\end{tabular}
	\end{table*}

	\begin{table*}
		\centering
		\caption{\label{tab7}%
			The TSS and BSS scores for $\geq$M-class flare prediction using the Transformer on the testing set with three distinct combinations of magnetic field parameters.} 
		\begin{tabular}{cccc}
			\hline
			\multirow{2}{*}{Feature}                                         & \multirow{2}{*}{Model}       & \multicolumn{2}{c}{Metric} \\ \cline{3-4} 
			&                              & TSS          & BSS         \\ \hline
			The   combination of  LOS and vector magnetic   field parameters & \multirow{3}{*}{Transformer} & 0.814±0.029  & 0.452±0.095 \\
			8 vector   magnetic field parameters                             &                              & 0.719±0.090  & 0.403±0.098 \\
			2  LOS   magnetic field parameters                               &                              & 0.697±0.137  & 0.448±0.079 \\ \hline
		\end{tabular}
	\end{table*}
	
	\begin{table*}
		\centering
		\caption{\label{tab8}%
			The categorical and probabilistic prediction performance of deep learning and {CDR} models across 10 testing sets from the 10 CV datasets.}.
		\begin{tabular}{ccccc}
			\hline
			\multirow{2}{*}{Model} & \multicolumn{4}{c}{Metric}                             \\ \cline{2-5} 
			& TSS          & Recall      & FPR         & BSS         \\ \hline
			CNN                    & 0.543±0.101  & 0.737±0.137 & 0.194±0.105 & 0.242±0.148 \\
			CNN-BiLSTM             & 0.682±0.085  & 0.813±0.060 & 0.132±0.060 & 0.443±0.105 \\
			\citet{111}       & 0.768±0.072 & 0.904±0.072 & 0.136±0.042 & 0.513±0.063 \\
			Transformer-10         & 0.814±0.029 & 0.970±0.033 & 0.156±0.027 & 0.452±0.095 \\
			Transformer-2          & 0.697±0.137  & 0.782±0.136 & 0.085±0.028 & 0.448±0.079 \\
			{CDR-CNN}               & {0.495±0.153}  & {0.789±0.115} & {0.294±0.186} & {0.076±0.073 }\\
			{CDR-CNN-BiLSTM}         & {0.453±0.132}  & {0.736±0.145} & {0.138±0.132} & {0.017±0.122} \\
			{CDR-Transformer-10}     & {0.829±0.059 } & {0.934±0.051} & {0.106±0.038} & {0.489±0.120} \\
			{CDR-Transformer-2}      & {0.714±0.062 } & {0.924±0.056} & {0.210±0.051} & {0.470±0.070} \\ \hline
		\end{tabular}
	\end{table*}

\begin{table*}
	\caption{{The paired t-test results between the CDR-Transformer and Transformer models.}}
	\label{tab9}
	\begin{tabular}{lll}
		\hline
		\multicolumn{1}{c}{\multirow{2}{*}{Models}} & \multicolumn{2}{c}{p-value} \\ \cline{2-3} 
		\multicolumn{1}{c}{}                                 & TSS          & BSS          \\ \hline
		CDR-Transformer-10 and   Transformer-10              & 0.262       & 0.094       \\
		CDR-Transformer-2 and   Transformer-2                & 0.366        & 0.431        \\ \hline
	\end{tabular}
\end{table*}

\begin{table*}
	\centering
	\captionsetup{justification=centering}
	\caption{{The effects of applying a perturbation (step size=1) to the TP reward on the TSS and BSS scores of the CDR-Transformer-10 model.}}
	\label{tab10}
	\begin{tabular}{cccccc}
		\hline
		TP   reward & TN   reward & FP   reward  & FN   reward  & TSS                  & BSS                  \\ \hline
		5           & 4           & -20          & -15          & 0.803±0.048          & 0.461±0.172          \\
		6           & 4           & -20          & -15          & 0.811±0.062          & 0.467±0.160          \\
		7           & 4           & -20          & -15          & 0.800±0.064          & 0.475±0.148          \\
		8           & 4           & -20          & -15          & 0.785±0.076          & 0.481±0.136          \\
		9           & 4           & -20          & -15          & 0.801±0.082          & 0.482±0.136          \\
		\textbf{10} & \textbf{4}  & \textbf{-20} & \textbf{-15} & \textbf{0.829±0.059} & \textbf{0.489±0.120} \\
		11          & 4           & -20          & -15          & 0.811±0.077          & 0.487±0.107          \\
		12          & 4           & -20          & -15          & 0.800±0.067          & 0.490±0.096          \\
		13          & 4           & -20          & -15          & 0.784±0.079          & 0.496±0.092          \\
		14          & 4           & -20          & -15          & 0.754±0.096          & 0.499±0.092          \\
		15          & 4           & -20          & -15          & 0.793±0.082          & 0.492±0.119          \\ \hline
	\end{tabular}
\end{table*}

\begin{table*}
	\centering
	\captionsetup{justification=centering}
	\caption{{The effects of applying a perturbation (step size=1) to the TN reward on the TSS and BSS scores of the CDR-Transformer-10 model.}}
	\label{tab11}
	\begin{tabular}{cccccc}
		\hline
		TP   reward & TN   reward & FP   reward  & FN   reward  & TSS                  & BSS                  \\ \hline
		10          & 1           & -20          & -15          & 0.797±0.067          & 0.492±0.119          \\
		10          & 2           & -20          & -15          & 0.826±0.061          & 0.492±0.117          \\
		10          & 3           & -20          & -15          & 0.815±0.062          & 0.489±0.119          \\
		\textbf{10} & \textbf{4}  & \textbf{-20} & \textbf{-15} & \textbf{0.829±0.059} & \textbf{0.489±0.120} \\
		10          & 5           & -20          & -15          & 0.826±0.063          & 0.488±0.121          \\
		10          & 6           & -20          & -15          & 0.821±0.070          & 0.486±0.123          \\
		10          & 7           & -20          & -15          & 0.821±0.070          & 0.488±0.116          \\ \hline
	\end{tabular}
\end{table*}

\begin{table*}
	\centering
	\captionsetup{justification=centering}
	\caption{{The effects of applying a perturbation (step size=1) to the FP reward on the TSS and BSS scores of the CDR-Transformer-10 model.}}
	\label{tab12}
	\begin{tabular}{cccccc}
		\hline
		TP   reward & TN   reward & FP   reward  & FN   reward  & TSS                  & BSS                  \\ \hline
		10          & 4           & -15          & -15          & 0.807±0.060          & 0.497±0.106          \\
		10          & 4           & -16          & -15          & 0.813±0.061          & 0.492±0.101          \\
		10          & 4           & -17          & -15          & 0.820±0.065          & 0.495±0.112          \\
		10          & 4           & -18          & -15          & 0.825±0.066          & 0.494±0.110          \\
		10          & 4           & -19          & -15          & 0.819±0.072          & 0.491±0.114          \\
		\textbf{10} & \textbf{4}  & \textbf{-20} & \textbf{-15} & \textbf{0.829±0.059} & \textbf{0.489±0.120} \\
		10          & 4           & -21          & -15          & 0.824±0.057          & 0.487±0.124          \\
		10          & 4           & -22          & -15          & 0.808±0.058          & 0.484±0.126          \\
		10          & 4           & -23          & -15          & 0.810±0.061          & 0.486±0.122          \\
		10          & 4           & -24          & -15          & 0.809±0.066          & 0.483±0.126          \\
		10          & 4           & -25          & -15          & 0.816±0.069          & 0.481±0.128          \\ \hline
	\end{tabular}
\end{table*}
	
	\begin{table*}
		\centering
		\captionsetup{justification=centering}
		\caption{{The effects of applying a perturbation (step size=1) to the FN reward on the TSS and BSS scores of the CDR-Transformer-10 model.}}
		\label{tab13}
		\begin{tabular}{cccccc}
			\hline
			TP reward   & TN   reward & FP   reward  & FN   reward  & TSS                  & BSS                  \\ \hline
			10          & 4           & -20          & -10          & 0.801±0.072          & 0.498±0.096          \\
			10          & 4           & -20          & -11          & 0.819±0.051          & 0.483±0.122          \\
			10          & 4           & -20          & -12          & 0.807±0.056          & 0.493±0.110          \\
			10          & 4           & -20          & -13          & 0.813±0.061          & 0.485±0.107          \\
			10          & 4           & -20          & -14          & 0.813±0.070          & 0.486±0.113          \\
			\textbf{10} & \textbf{4}  & \textbf{-20} & \textbf{-15} & \textbf{0.829±0.059} & \textbf{0.489±0.120} \\
			10          & 4           & -20          & -16          & 0.801±0.050          & 0.490±0.123          \\
			10          & 4           & -20          & -17          & 0.800±0.072          & 0.493±0.124          \\
			10          & 4           & -20          & -18          & 0.802±0.056          & 0.489±0.123          \\
			10          & 4           & -20          & -19          & 0.806±0.070          & 0.487±0.130          \\
			10          & 4           & -20          & -20          & 0.811±0.063          & 0.485±0.135          \\ \hline
		\end{tabular}
	\end{table*}

	\begin{table*}
		\centering
		\caption{The performance results for {CDR-Transformer-10}, Transformer-10, iTransformer, and NASA/CCMC at the probability thresholds corresponding to their respective optimal TSS.}\label{tab14}%
		\begin{threeparttable}
			
			\begin{tabular}{llccc}
				\hline
				Data & Model & TSS & Recall & FPR \\
				\hline
				\multirow{4}{*}{Filtered testing dataset} & NASA/CCMC & 0.853 & 1 & 0.147 \\
				& iTransformer & 0.906 $\pm$ 0.043 & 1.000 $\pm$ 0.000 & 0.094 $\pm$ 0.043 \\
				& Transformer & 0.933 $\pm$ 0.057 & 0.980 $\pm$ 0.060 & 0.047 $\pm$ 0.020 \\
				& {CDR-Transformer} & {0.968 $\pm$ 0.016}  & 1.000 $\pm$ 0.000 & {0.032 $\pm$ 0.016} \\
				\hline
				\multirow{4}{*}{Original testing dataset} & NASA/CCMC & 0.633 & 0.706 & 0.072 \\
				& iTransformer & 0.668 $\pm$ 0.060 & 1.000 $\pm$ 0.000 & 0.332 $\pm$ 0.060 \\
				& Transformer & 0.652 $\pm$ 0.060 & 0.859 $\pm$ 0.088 & 0.207 $\pm$ 0.063 \\
				& {CDR-Transformer} & {0.635 $\pm$ 0.045} & {0.818 $\pm$ 0.081} & {0.183 $\pm$ 0.051} \\
				\hline
			\end{tabular}
			\begin{tablenotes}
				\footnotesize
				\item \textbf{Note:} The results of iTransformer are from \citet{111}.
			\end{tablenotes}
		\end{threeparttable}
	\end{table*}

	\bsp	
	\label{lastpage}
\end{document}